\documentclass[11pt,a4paper]{article}
\usepackage[hyperref]{eacl2021}
\aclfinalcopy

\usepackage{subcaption}
\usepackage{times}
\usepackage{latexsym}

\usepackage{svg}
\usepackage{graphicx}
\usepackage[inline]{enumitem}

\usepackage{microtype}
\usepackage{multirow}

\usepackage[font=small,labelfont=bf]{caption}
\usepackage{amsmath}
\usepackage{amsfonts}
\usepackage{amssymb}
\usepackage{cleveref}
\crefformat{section}{\S#2#1#3} 
\crefformat{subsection}{\S#2#1#3}
\crefformat{subsubsection}{\S#2#1#3}
\crefname{figure}{Fig.}{Figs.}

\usepackage{makecell}
\usepackage{xcolor}
\usepackage{placeins}

\usepackage{array}
\newcolumntype{U}{>{\centering\arraybackslash}m{1.8cm}}
\newcolumntype{V}{>{\centering\arraybackslash}m{1.2cm}}
\newcolumntype{Z}{>{\centering\arraybackslash}m{0.9cm}}
\newcolumntype{W}{>{\centering\arraybackslash}m{1in}}

\usepackage{floatrow}
\DeclareFloatFont{small}{\small}
\newcommand\Tstrut{\rule{0pt}{2.0ex}}         
\newcommand{\norm}[1]{\left\lVert#1\right\rVert}

\makeatletter
\newcommand\footnoteref[1]{\protected@xdef\@thefnmark{\ref{#1}}\@footnotemark}
\makeatother

\makeatletter
\newcommand*\dotproduct{\mathpalette\dotproduct@{.5}}
\newcommand*\dotproduct@[2]{\mathbin{\vcenter{\hbox{\scalebox{#2}{$\m@th#1\bullet$}}}}}
\makeatother

\title{{L}ocality {P}reserving {L}oss: {N}eighbors that \textit{{L}ive} together, \textit{{A}lign} together}

\author{Ashwinkumar Ganesan, Francis Ferraro, Tim Oates \\
  Dept. Of Computer Science \& Electrical Engineering (CSEE), \\
  University Of Maryland Baltimore County (UMBC), \\
  MD, USA - 21250 \\
  {\tt gashwin1@umbc.edu, ferraro@umbc.edu, oates@cs.umbc.edu} \\}

\date{}

\begin{document}
\maketitle
\begin{abstract}
We present a \textbf{locality preserving loss} (LPL) that improves the alignment between vector space embeddings while separating uncorrelated representations. Given two pretrained embedding manifolds, LPL optimizes a model to project an embedding and maintain its local neighborhood while aligning one manifold to another. This reduces the overall size of the dataset required to align the two in tasks such as crosslingual word alignment. We show that the LPL-based alignment between input vector spaces acts as a regularizer, leading to better and consistent accuracy than the baseline, especially when the size of the training set is small. We demonstrate the effectiveness of LPL-optimized alignment on semantic text similarity (STS), natural language inference (SNLI), multi-genre language inference (MNLI) and cross-lingual word alignment (CLA) showing consistent improvements, finding up to $16$\% improvement over our baseline in lower resource settings.\footnote{Code Source: \href{https://github.com/codehacken/locality-preservation}{https://github.com/codehacken/locality-preservation}}

\end{abstract}

\section{Introduction}
\label{sec:Introduction}
\begin{figure*}[ht]
    \centering
    \includegraphics[scale=0.43]{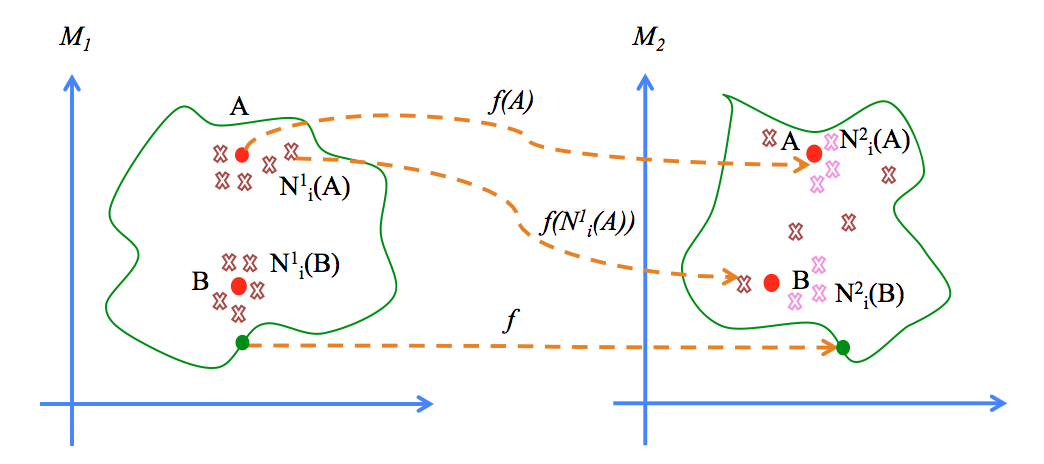}
    \includegraphics[scale=0.43]{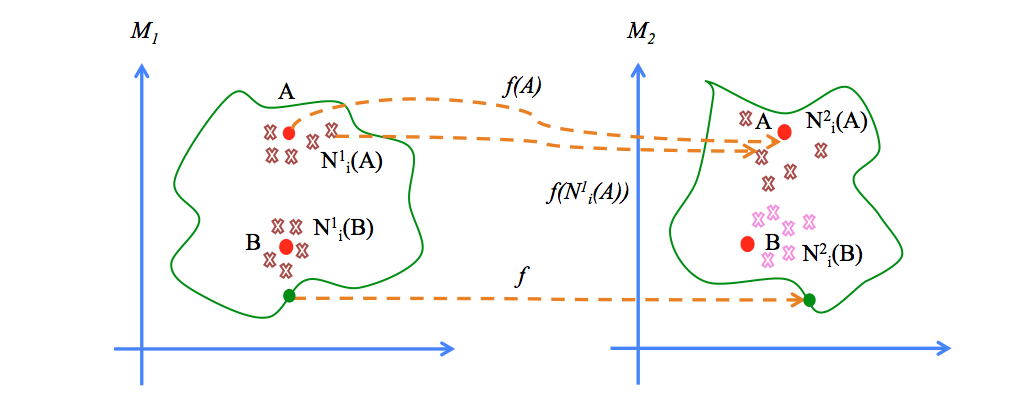}
\caption{\textbf{Quality of alignment with different types of losses.} $A$, $B$ are two words in two word embedding manifolds $M_1$ and $M_2$. $f$ is the manifold alignment function while $N_i^1(A)$ and $N_i^1(B)$ are their respective neighbors in manifold $M_1$. $N_i^2(A)$ and $N_i^2(B)$ are their neighbors in manifold $M_2$. Figure (a) shows the alignment when trained with a MSE loss. The neighbors are distributed across the manifold due to overfitting. (b) shows alignment with a locality preserving loss (LPL) that reconstructs the original manifold in the target domain $M_2$ maintaining its local structure.}
    \label{fig:align_mse_lpp}
\end{figure*}


Over the last few years, vector space representations of words and sentences, extracted from encoders trained on a large text corpus, are primary components to model any natural language processing (NLP) task, especially while using neural or deep learning methods. Neural NLP models can be initialized with pretrained word embeddings learned using \textit{word2vec} \citep{mikolov2013distributed} or \textit{GloVe} \citep{pennington2014glove} and fine-tuned sentence encoders like BERT \citep{devlin2018bert} or RoBERTa \citep{liu2019roberta}. They show state-of-the-art performance in a number of tasks from part-of-speech tagging, named entity recognition, and machine translation to measuring textual similarity \citep{wang2018glue}. BERT's success has spawned research dedicated to understanding \citep{rogers2020primer} and reducing parameters in transformer architectures \citep{lan2019albert}. Despite these successes, supervised transfer learning is not a panacea. Models based on pretrained word embeddings (for bilingual induction) or BERT-based models require a large parallel corpus to train on. Can we reduce the number of training samples even further? 

We approach this problem by proposing a framework that exploits the inherent relationship between word or sentence representations in their \textit{pretrained manifolds}. These relationships help train the model with fewer samples, since each training sample represents a group of instances.


We consider three types of tasks: 
\begin{enumerate*}[label={(\arabic*)}]
    \item a regression task such as semantic text similarity;
    \item a classification task (e.g., natural language inference (NLI)); and
    \item vector space alignment, where the purpose is to learn a mapping between two independently trained embeddings (e.g., crosslingual word alignment)
\end{enumerate*}
Learning bilingual word embedding models alleviates low resource problems by aligning embeddings from a source language that is rich in available text to a target language with a small corpus with limited vocabulary.\footnote{\textit{Low resource} has two interpretations i.e. one where corpus size to generate unsupervised pretrained embeddings is small and the other where the parallel corpus for alignment is minimal. We experiment with the later here.} Largely, recent work focuses on learning a linear mapping to align two embedding spaces by minimizing the mean squared error (MSE) between embeddings of words projected from the source domain and their counterparts in the target domain \citep{mikolov2013distributed,ruder2017survey}. Minimizing MSE is useful when a large set of translated words (between source and target languages) is provided, but the mapping overfits when the parallel corpus is small or may require non-linear transformations \citep{sogaard2018limitations}. In order to reduce overfitting and improve word alignment, we propose an auxiliary loss function called locality preserving loss (LPL) that trains the model to align two sets of word embeddings while maintaining the local neighborhood structure around words in the source domain.

With classification and regression tasks where there are two inputs (e.g., NLI and STS-B), we show how the alignment between the two input subspace acts as regularizer, improving the model's accuracy on the task with MSE alone and when MSE and LPL are combined together. LPL achieves this by augmenting existing text $\leftrightarrow$ label pairs with pseudo-pairs constructed from their neighbors.

Specifically, our main contributions are:

\begin{itemize}
\setlength{\itemsep}{0.20pt}
    \item We propose a loss function called \textbf{locality preserving loss (LPL)} to improve vector space alignment and show how the \textbf{alignment acts as a regularizer} while performing language inference and semantic text similarity. LPL improves correlation or accuracy of linear and non-linear mapping (deep networks) while exploiting the inherent geometries of existing pretrained embedding manifolds to optimize an alignment model (Table \ref{tab:exp_alignment_performance}, Figure \ref{fig:exp_stsb_pearson}).
    
    \item We show LPL is flexible and can be optimized with SGD. Hence, it can be applied to both deep neural networks and linear transformations. In contrast, previous cross-lingual word alignment models that are a linear map between source and target language, are learned using singular value decomposition.
        
    \item We show an increase in correlation on semantic text similarity (STS-B) and accuracy on SNLI in comparison with the baseline when the models are trained with small datasets. Training with LPL shows up to $16.38$\% ($90.1$\% relative), on SNLI up to $8.9$\% ($19.3$\% relative) improvement when trained with just $1000$ samples ($0.002$\% of the dataset). We train a crosslingual word alignment model giving up to 4.1\% (13.8\% relative) improvement in comparison to a MSE optimized mapping while reducing the size of the parallel corpus required to train the mapping by $40$\% ($3$K out of $5$K pairs).
\end{itemize}

    
    

\section{Background \& Related Work}

\label{sec:RelatedWork}
\subsection{Dimensionality Reduction}
\label{sec:related_work_dr}
Manifold learning methods represent these high dimensional datapoints in a lower dimensional space by extracting important features from the data, making it easier to cluster and search for similar data points. The methods are broadly categorized into linear, such as Principal Component Analysis (PCA), and non-linear algorithms. Non-linear methods include multi-dimensional scaling~\cite[MDS]{cox2000multidimensional}, locally linear embedding~ \cite[LLE]{roweis2000nonlinear} and Laplacian eigenmaps~\cite[LE]{belkin2002laplacian}. \citet{he2004locality} compute the Euclidean distance between points to construct an adjacency graph and create a linear map that preserves the neighborhood structure of each point in the manifold. Another popular tool to learn manifolds is an autoencoder where a self-reconstruction loss is used to train a neural network \citep{rumelhart1985learning}. \citet{vincent2008extracting} design an autoencoder that is robust to noise by training it with a noisy input and then reconstructing the original noise-free input. 

In locally linear embedding (LLE) \citep{roweis2000nonlinear}, the datapoints are assumed to have a linear relation with their neighbors. To project each point, first a reconstruction loss is utilized to learn the linear relation between a point and their $k$ neighbors. Then, the linear relation is used to learn the embeddings in the reduced dimension.\footnote{See \Cref{sec:app_lle_desc} for an in-depth explanation of LLE.}


\citet{wang2014generalized} extend autoencoders by modifying the reconstruction loss to use nearest neighbors of data points, leveraging neighborhood relationships between datapoints from non-linear dimension reduction methods like LLE and Laplacian Eigenmaps. 



\subsection{Manifold Alignment}
\citet{benaim2017one} utilize a GAN to learn a unidirectional mapping. The total loss applied to train the generator is a combination of different losses, namely, an adversarial loss, a cyclic constraint (inspired by \citet{zhu2017unpaired}), MSE and an additional distance constraint where the distance between the point and its neighbors in the source domain are maintained in the target domain. Similarly, \citet{conneau2017word} learn to translate words without any parallel data with a GAN that optimizes a cross domain similarity scale to resolve the hubness problem \citep{dinu2014improving}.

These methods are the foundation to learn a mapping between two lower dimensional spaces (manifold alignment, Figure \ref{fig:align_mse_lpp}). \Citet{wang2011manifold} propose a manifold alignment method that preserves the local similarity between points in the manifold being transformed and the correspondence between points that are common to both manifolds. \Citet{boucher2015aligning} replace the manifold alignment algorithm that uses the nearest neighbor graph with a low rank alignment. \citet{cui2014generalized} align two manifolds without any pairwise data (unsupervised) by assuming the structure of the lower dimension manifolds are similar. 

Our work is similar to \citet{bollegala2017think} where the meta-embedding (a common embedding space) for different vector representations is generated using a locally linear embedding (LLE) which preserves the locality. One drawback, though, is that LLE does not learn a singular functional mapping between the source and target vector spaces. A linear mapping between a word and its neighbor is learned for each new word. Hence, the meta-embedding must be retrained every time new words are added to the vocabulary. \citet{nakashole2018norma} propose NORMA that uses neighborhood sensitive maps where the neighbors are learned rather than extracted from the existing embedding space. Similar to NORMA, LPL uses a modified locally linear representation of each embedding but, unlike it, LPL uses actual nearest neighbors in order to learn an embedding. This is important as NNs may not be present in annotated parallel corpus. \textbf{Hence, using NNs of annotated pairs in the corpus in LPL expands the size of the training dataset.} LPL is optimized with gradient descent and can be easily added to a deep neural network (as seen in \cref{subsec:align_as_reg}).

\section{Incorporating Locality Preservation into Task-based Learning}
\label{sec:ProblemDefinition}
In this section, we describe locality preserving loss, the assumption underlying the loss function and objective functions (\cref{eq:alignment_lp_loss}, \cref{eq:alignment_lle}) that are optimized. The cumulative loss function while training the model is defined in \cref{eq:alignment_total_loss_ph2}.

\subsection{Locality Preservation Criteria}
The locality preserving loss (LPL, \cref{eq:alignment_lp_loss}) is based on an important assumption about the source manifold: for a pre-defined neighborhood of $k$ points ($k$ is chosen manually) in the source embedding space we assume points are ``close" to a given point such that it can be reconstructed using a linear map of its neighbors. This assumption is similar to that made in locally linear embedding \citep{roweis2000nonlinear}. 

\subsection{Preliminaries}
\label{subsec:ma_ProblemDefinition}

As individual embeddings can represent words or sentences, we call each individual embedding a \textit{unit}. Consider two manifolds---$M^s\in\mathbb{R}^{n\times d}$ (source domain) and $M^t\in\mathbb{R}^{m\times d}$ (target domain)---that are vector space representations of units within each domain. We do not make assumptions on the methods used to learn each manifold; they may be different. We also do not assume they share a common lexical vocabulary. For example, $M^s$ can be created using a standard distributed representation method like word2vec \citep{mikolov2013distributed} and consists of English word embeddings while $M^t$ is created using GloVe \citep{pennington2014glove} and contains Italian embeddings. Let $V^s$ and $V^t$ be the respective vocabularies (collection of units) of the two manifolds. Hence $V^s = \{w^s_1\ ...\ w^s_n\}$ and $V^t = \{w^t_1\ ..\ w^t_m\}$ are sets of units in each vocabulary of size $n$ and $m$. The distributed representations of the units in each manifold are $M^s = \{m^s_1\ ...\ m^s_n\}$ and $M^t = \{m^t_1\ ...\ m^t_m\}$.

While we do not assume that $V^t$ and $V^s$ must have common items, we do assume that there is some set of unit pairs that are connected by some consistent relationship. Let $V^p = \{w^p_1\ ...\ w^p_c\}$ be the set of the unit pairs; we consider $V^p$ a supervised training set (though it could be weakly supervised, e.g., derived from a parallel corpus). For example, in crosslingual word alignment this consistent relationship is whether one word can be translated as another; in natural language inference, the relationship is whether one sentence entails the other (the second must logically follow from the first). We assume this common set $V^p$ is much smaller than the individual vocabularies ($c << m$ and $c << n$). The mapping (manifold alignment) function is $f$.

In this paper, we experiment with three types of tasks: cross-lingual word alignment (mapping),  natural language inference (classification), and semantic text similarity (regression). In cross-lingual word alignment, $V^s$ and $V^t$ represent the source and target vocabularies, $V^p$ is a bilingual dictionary, and $M^t$ and $M^s$ are the target and source manifolds. $f$ with $\theta_f$ parameters is a linear projection with a single weight matrix $W$. For NLI, $V^t$ and $V^s$ are target and source sentences with $M^t$ and $M^s$ being their manifolds. $f$ is a $2$-layer FFN.

\subsection{Locality Preserving Loss (LPL)}
We use a mapping function $f: M^s \rightarrow M^t$ to align manifold $M^s$ to $M^t$. The exact structure of $f$ is task-specific: for example, in our experiments $f$ is a linear function for crosslingual word alignment and it is a $2$-layer neural network (non-linear mapping) for NLI. The mapping is optimized using three loss functions: an orthogonal transform \citep{xing2015normalized} $\mathcal{L}_{\textrm{ortho}}$ (constrain $W^{-1} = W^T$); mean squared error $\mathcal{L}_{\textrm{mse}}$ (eq. \ref{eqn:l_mse}); and locality preserving loss (LPL) $\mathcal{L}_{\textrm{lpl}}$ (eq. \ref{eq:alignment_lp_loss}).

The standard loss function to align two manifolds is mean squared error (MSE) \citep{ruder2017survey, artetxe2016learning},
\begin{equation}
\mathcal{L}_{\textrm{mse}} = \sum_{i \in V^p} \mathcal{L}_{\textrm{mse}}^i = \sum_{i \in V^p}\overbrace{\left\|f(m^s_i) - m^t_i\right\|^2_2}^{\mathcal{L}_{\textrm{mse}}^i},
\label{eqn:l_mse}
\end{equation}
which minimizes the distance between the unit's representation in $M_t$ (the target manifold) and projected vector from $M_s$. The function $f(m^s_i)$ has learnable parameters $\theta_f$.
MSE can lead to an optimal alignment when there is a large number of units in the parallel corpus to train the mapping between the two manifolds \citep{ruder2017survey}. However, when the parallel corpus $V^p$ is small, the mapping is prone to overfitting \citep{glavas2019properly}.

Locality preserving loss (LPL: eq. \ref{eq:alignment_lp_loss}) optimizes the mapping $f$ to project a unit together with its neighbors. For a small neighborhood of $k$ units, the source representation of unit $w^s_i$ is assumed to be a linear combination of its source neighbors. We represent this small neighborhood (of the source embedding $m^s_i$ of word $w^s_i$) with $N_k(m^s_i)$, and we compute the local linear reconstruction using $W_{ij}$, a learned weight associated with each word in the neighborhood of the current word, $N_k(m^s_i)$. %
LPL requires that the projected source embedding $f(m^s_i)$ be a weighted average of all the projected vectors of its neighbors $f(m^s_j)$. %
Formally, for a particular common item $i$, LPL at $i$ minimizes
\begin{equation}
\mathcal{L}_{\textrm{lpl}}^i =\norm{m^t_i - \sum_{m^s_j \in N_k(m^s_i)}{ W_{ij} f(m^s_j)}}^2
\label{eq:alignment_lp_loss}
\end{equation}
with $\mathcal{L}_{\textrm{lpl}} = \sum_{m^s_i, m^t_i \in V^s}\mathcal{L}_{\textrm{lpl}}^i$. %
Intuitively, $W$ represents the relation between a word and its neighbors in the source domain. We learn it by minimizing the LLE-inspired loss. For a common $i$ this is
\begin{equation}
\mathcal{L}_{\textrm{lle}}^i =  \norm{m^s_i - \sum_{m^s_j \in N_k(m^s_i)}{ W_{ij} m^s_j}}^2
\label{eq:alignment_lle}
\end{equation}
with $\mathcal{L}_{\textrm{lle}} =  \sum_{m^s_i \in V^p} \mathcal{L}_{\textrm{lle}}^i$. %
The weights $W$ are subject to the constraint $\sum{W_{ij} = 1}$, making the projected embeddings invariant to scaling~\citep{roweis2000nonlinear}. We can formalize this with an objective $\mathcal{L}_{\textrm{ortho}} = WW^\intercal - I$. LPL reduces overfitting because the mapping function $f$ does not simply learn the mapping between unit embeddings in the parallel corpus: it also optimizes for a projection of the unit's neighbors that are not part of the parallel corpus---effectively expanding the size of the training set by the factor $k$.

\subsection{Model Training with LPL}
\label{subsubsec:lpa}
The total supervised loss becomes:
\begin{multline}
\mathcal{L}_{\textrm{sup}} = \mathcal{L}_{\textrm{mse}}(\theta_f) + \beta*\mathcal{L}_{\textrm{lpl}}(\theta_f, W) + \mathcal{L}_{\textrm{lle}}(W) \\ + \mathcal{L}_{\textrm{ortho}}(\theta_f)
\label{eq:alignment_total_loss_ph2}
\end{multline}
We introduce a constant $\beta$ to allow control over the contribution of LPL to the total loss. Although we minimize total loss \eqref{eq:alignment_total_loss_ph2}, shown explicitly with variable dependence, the optimization can be unstable as there are two sets of independent parameters $W$ and $\theta_f$ representing different relationships between datapoints. To reduce the instability, we split the training into two phases. In the first phase, $W$ is learned by minimizing $\mathcal{L}_{\textrm{lle}}$ alone and the weights are frozen. Then, $\mathcal{L}_{\textrm{mse}}$ and $\mathcal{L}_{\textrm{lpl}}$ are minimized while keeping $W$ fixed.


One key difference between our work and \citet{artetxe2016learning} is that they optimize the mapping function by taking the singular vector decomposition (SVD) of the squared loss while we use gradient descent to find optimal values of $\theta_f$. As our experimental results show, while both can be empirically advantageous, our work allows LPL to be easily added as just another term in the loss function. With the exception of the alternating optimization of $W$, our approach does not need special optimization updates to be derived. \textbf{Euclidean distance} between embeddings is used to find NNs.

\subsection{Alignment as Regularization}
\label{subsec:align_as_reg}
\begin{figure}[t]
    \centering
    \includegraphics[scale=0.205]{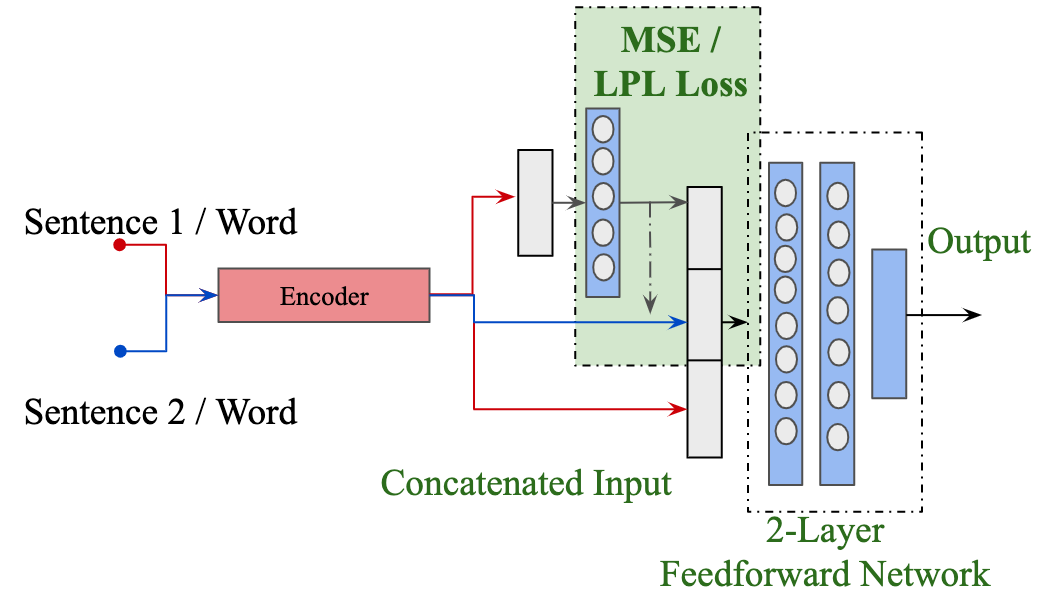}
\caption{\textbf{Use of alignment loss for the NLI and STS-B task.} The pipeline consists of a $2$-layer FFN used to classify sentence pairs. The output layer is of size $3$ to classify the input into \textit{entailment}, \textit{contradiction} and \textit{neutral}. It has a size of $1$ to generate a continuous value between $0$ and $1$ for STS-B. The premise / sentence 1 and hypothesis / sentence 2 subspaces are aligned using a MSE and LPL loss that are then added as a concatenated input to train the classifier / regressor. A $\delta$ hyperparameter configured for each label provides the ability to perform \textit{alignment} for \textit{entailment} and \textit{contradiction} while performing \textit{divergence} for \textit{neutral} input pairs.}
    \label{fig:exp_align_network}
\end{figure}

MSE and LPL can be used to align two vector spaces: in particular, we show that the objectives can align two subspaces in the \textit{same} manifold. When combined with cross entropy loss in a classification task, this subspace alignment effectively acts as a regularizer. Figure \ref{fig:exp_align_network} shows an example architecture where alignment is used as a regularizer for the NLI task. The architecture contains a two layer FFN used to perform language inference, i.e., to predict if the given sentence pairs are \textit{entailed}, \textit{contradictory} or \textit{neutral}. The input to the network is a pair of sentence vectors. The initial representations are generated from any sentence/language encoder, e.g., from BERT. The source/sentence$1$/premise embeddings are first projected to the hypothesis space. The projected vector is then concatenated with the original pair of embeddings and given as input to the network. The alignment losses (MSE and LPL) are computed between the projected premise and original hypothesis embeddings. If the baseline network is optimized with cross entropy (CE) loss to predict label $y_i$, the total loss becomes:
\begin{equation}
\mathcal{L}_{\textrm{total}} = \gamma \sum_i{\delta_{y_i} (\mathcal{L}_{\textrm{mse}}^{i} + \mathcal{L}_{\textrm{lpl}}^{i} + \mathcal{L}_{\textrm{lle}}^i(W)) + \textit{CE}_{y_i}}
\label{eq:alignment_regularizer}
\end{equation}
where $\gamma$ is a hyperparameter that controls the impact of the loss (learning rate). Thus, the loss (\cref{eq:alignment_regularizer}) is an extension of \cref{eq:alignment_total_loss_ph2} for a classification task but without $\mathcal{L}_{\textrm{ortho}}$, which is not applied as $f$ is a $2$-layer FFN (non-linear mapping) and the $W W^{\intercal} = I$ constraint for each layer's weights cannot be guaranteed. The alignment loss becomes a vehicle to bias the model based upon our knowledge of the task, forcing a specific behavior on the network. The behavior can be controlled with $\delta$, which can be a positive or negative value specific to each label. A positive $\delta$ optimizes the network to align the embeddings while a negative $\delta$ is a divergence loss. In NLI we assign a constant scalar to all samples with a specific label (i.e., $100$ for entailment, $1.0$ for contradiction and $-5.0$ for neutral). The scalars were set when optimizing network hyper-parameters.  As the optimizer minimizes the loss, a divergence loss tends to $-\infty$; in practice, we clip the negative loss value at $-1$. 

\section{Experiment Results \& Analysis}
\label{sec:experiments}

We demonstrate the effectiveness of the locality preserving alignment on three types of tasks: semantic text similarity (regression), natural language inference (text classification) and crosslingual word alignment (mapping / regression). In order to compute local neighborhoods, as needed for, e.g., \eqref{eq:alignment_lp_loss}, we build a standard KD-Tree and find the nearest neighbors using Euclidean distance. 

\subsection{Semantic Text Similarity (STS)}
\label{subsec:exp_sts}
\begin{figure}[t]
    \centering
    \includegraphics[width=\textwidth]{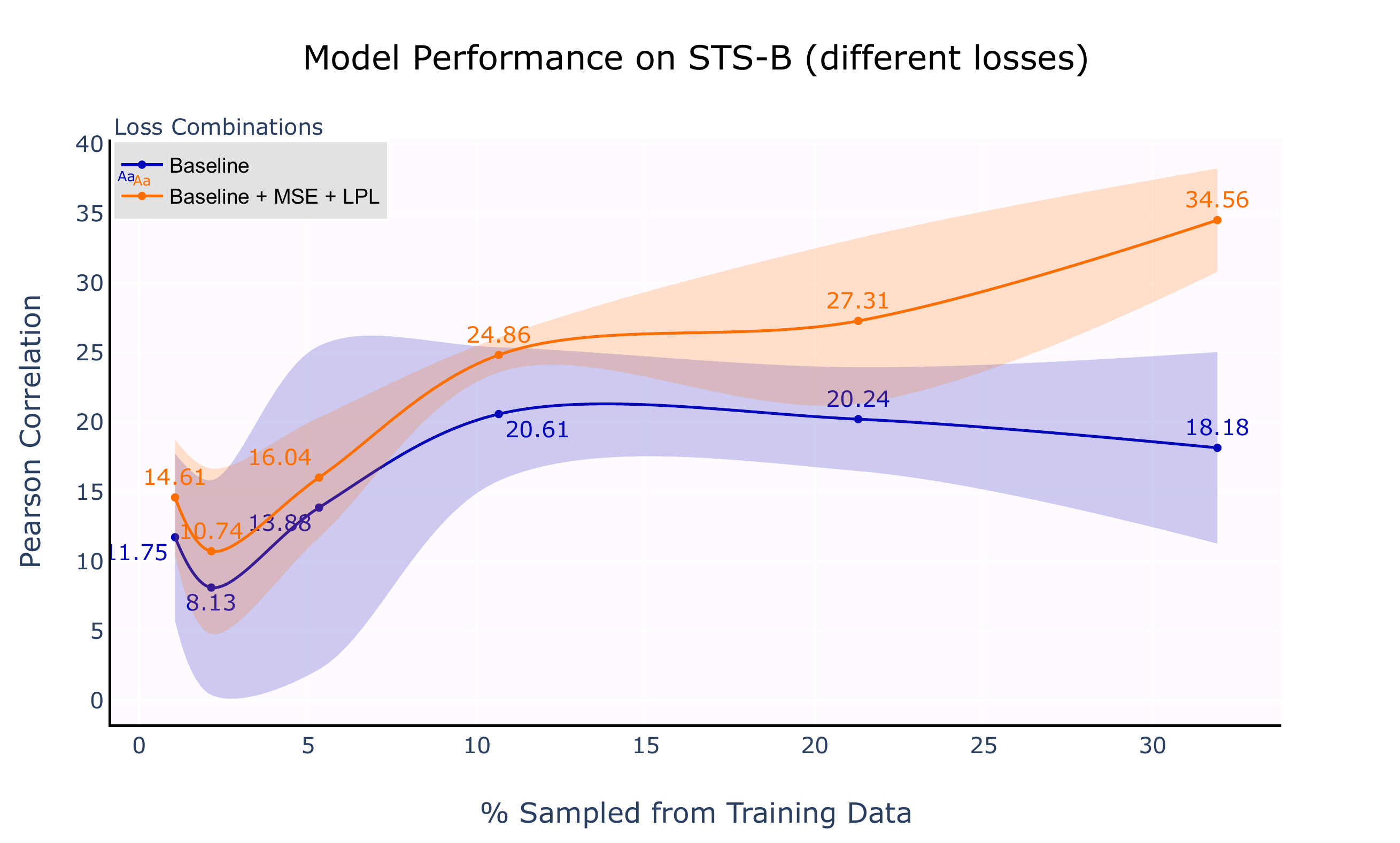}
\caption{Pearson correlation on semantic text similarity with STS-B dataset. Locality Preserving Loss (LPL) improves sentence pair similarity in all data sizes from small ($100$ samples or $2$\% of total data) to $50\%$ of the dataset.}
    \label{fig:exp_stsb_pearson}
\end{figure}
In semantic text similarity (STS), we use the STS-B dataset \citep{cer2017semeval} that is widely utilized as a part of the GLUE benchmark \citep{wang2018glue}. The baseline model is the same as Siamese-BERT \citep{reimers2019sentence} where sentence embeddings are extracted individually using a BERT model \citep{devlin2018bert}. An attached connected feedforward network (FFN) is trained to predict a sentence similarity score between $0$ and $1$ (optimized with squared error loss).\footnote{Appendix A.1 provides details about the dataset and FFN.}

To analyze the impact of various loss functions, the BERT model parameters are frozen so that sentence embeddings remain the same and only the parameters of the FFN are optimized. As described in \cref{subsubsec:lpa}, the baseline model is additionally trained with a \texttt{MSE} loss that aligns one sentence manifold to another ($m_1 \rightarrow m_2$) and the third model also trains with \texttt{LPL}. $\delta$ is set to mimic the normalized label thus generating the largest loss while aligning two sentences that are same. The contrastive loss separates the two sentence embeddings when they are dissimilar. The maximum margin is set to $0.1$.


In order to study the impact on model performance when the training data is small, we limit the sampled training data size to upto $50\%$ of the original (total dataset size is 5k).\textsuperscript{\ref{all_data_graph}} Figure \ref{fig:exp_stsb_pearson} shows the Pearson correlation between the baseline model and the same regularizer with \texttt{MSE} and \texttt{LPL} (eq. \ref{eq:alignment_total_loss_ph2}). As observed, the correlation of models trained with \texttt{LPL} is higher for every training dataset size. The relative increase in correlation is in the range of $10.83$ to $90.1$\%. We note here that the correlation cannot be compared with the original BERT model as we do not fine-tune the entire network but only the FFN in order to measure the improvements with the addition of \texttt{LPL}. 

\begin{table*}
    \centering
    \begin{subtable}[t]{.65\columnwidth}
    \resizebox{\columnwidth}{!}{
    \begin{tabular}{W | c | c | c | c | c | c}
        \hline
         \textbf{Method} & \textbf{Trans.} & \textbf{Optim.} & \textbf{EN-IT} & \textbf{EN-DE} & \textbf{EN-FI} & \textbf{EN-ES}\Tstrut \\ 
        \hline
        MSE, Train T$\rightarrow$S \citep{shigeto2015ridge} & S1, S2 & Linear & 41.53 & 43.07 & 31.04 & 33.73\Tstrut\\
        MSE \citep{artetxe2016learning} & S0, S2 & Linear & 39.27 & 41.87 & 30.62 & 31.40 \\
        MSE: IS \citep{smith2017offline} & S0, S2, S5 & Linear & 41.53 & 43.07 & 31.04 & 33.73 \\
        MSE: NN \citep{artetxe2018generalizing} & S0-S5 & Linear & 44.00 & 44.27 & 32.94 & 36.53 \\
        MSE: IS \citep{artetxe2018generalizing} & S0-S5 & Linear & \textbf{45.27} & 44.13 & 32.94 & \textbf{36.60} \\
        \hline
        MSE & S0, S2 & SGD & 39.67 & 45.47 & 29.42 & 35.3\Tstrut \\
        \hline
        \textbf{LPL+MSE: CSLS} & S0, S2 & SGD & 43.33 & \textbf{46.07} & \textbf{33.50} & 35.13\Tstrut \\
        \hline
    \end{tabular}
    }
  \caption{We compare our method (bottom row: LPL) on cross-lingual word alignment. In comparison to \citet{artetxe2018generalizing}, we use cross-domain similarity local scaling (CSLS) \citep{conneau2017word} to retrieve the translated word. The \textit{method} column lists different losses/methods used to learn the projection: \textit{NN} is nearest neighbor search while \textit{IS} is inverted softmax. Many mapping methods use additional transformation steps.}
  \label{tab:exp_alignment_performance}
  \end{subtable}
  ~
  \begin{subtable}[t]{.32\columnwidth}
    \centering
    \resizebox{\columnwidth}{!}{
    \small
    \begin{tabular}{Z | U | V}
        \hline
        \textbf{Trans.} & \textbf{Desc.} & \textbf{Backprop?}\\ 
        \hline
        S0 & Embedding normalization (unit / center) & Yes\\\hline
        S1 &  Whitening & No\\\hline
        S2 &  Orthogonal Mapping & Yes\\\hline
        S3 &  Re-weighting & No\\\hline
        S4 &  De-Whitening & No\\\hline
        S5 &  Dimensionality Reduction & No\\
        \hline
    \end{tabular}
    }
  \caption{A map of various transformations that can be performed as described in \citet{artetxe2018generalizing}. We indicate which steps can easily be combined with backpropagation.}
  \label{tab:exp_alignment_steps_desc}
\end{subtable}
\caption{The accuracy of the locality preserving method. \Cref{tab:exp_alignment_performance} lists 6 high-performing supervised/semi-supervised baselines; \cref{tab:exp_alignment_steps_desc} lists the  transformations used in these methods and how easily those transformations can be used with back-propagation. Notice that our method uses transformations amenable with back-propagation. In \Cref{tab:exp_alignment_performance}, the first five baselines rely on algebraic updates while our method works nicely with SGD: we include the sixth row (MSE via SGD) to illustrate the comparative performance gain we obtain.}
\label{tab:exp_alignment}
\end{table*}

\subsection{Natural Language Inference}
\label{subsec:exp_nli}
\begin{figure}[t]
    \centering
    \includegraphics[width=\textwidth]{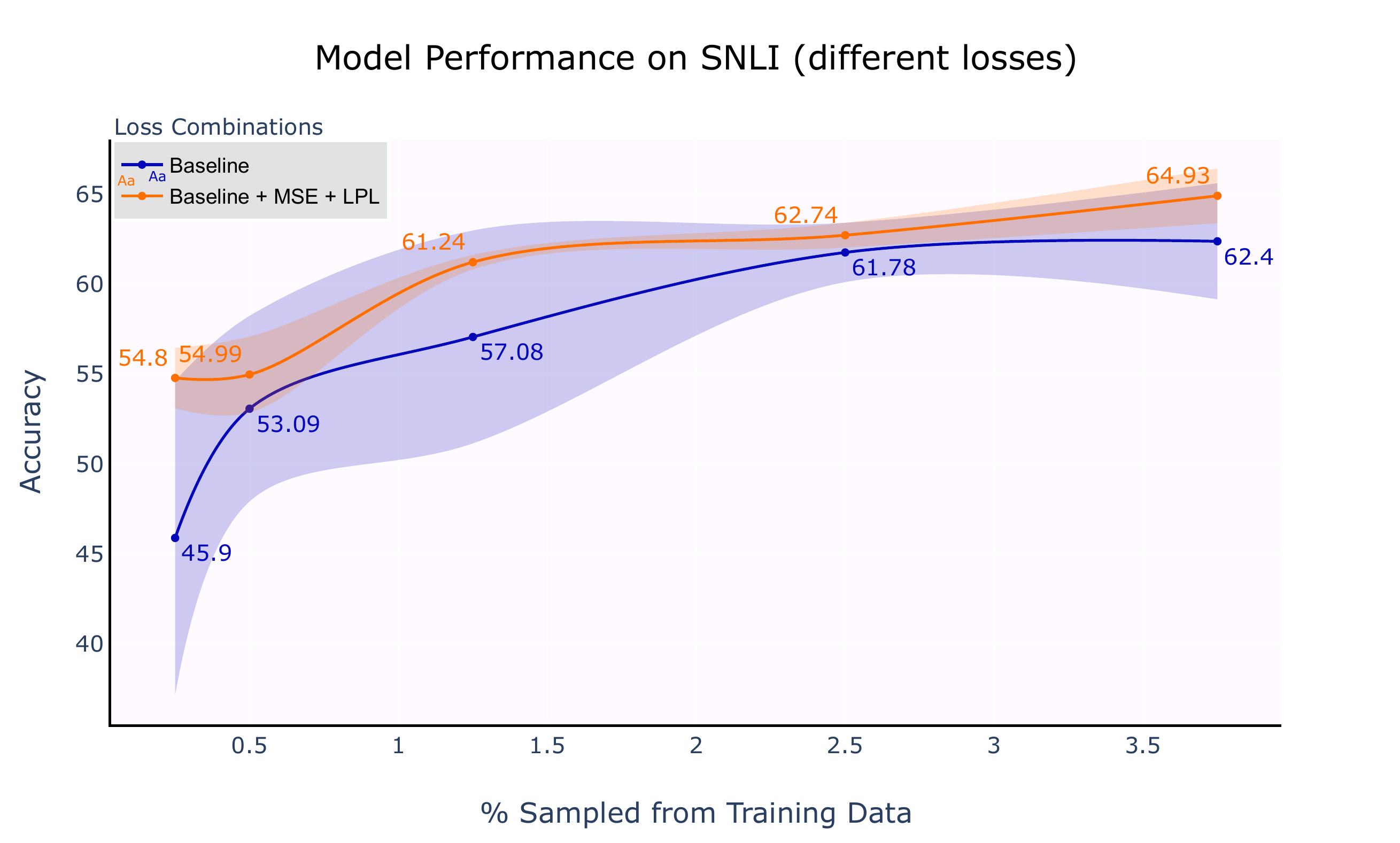}
\caption{Accuracy of alignment regularization on SNLI. The graph shows the accuracy, averaged across 3 runs, for differing size of training samples up to $5\%$ of the training dataset only (total: $500$K).}
    \label{fig:exp_snli_perf}
\end{figure}

\begin{figure}[t]
    \centering
    \includegraphics[width=\textwidth]{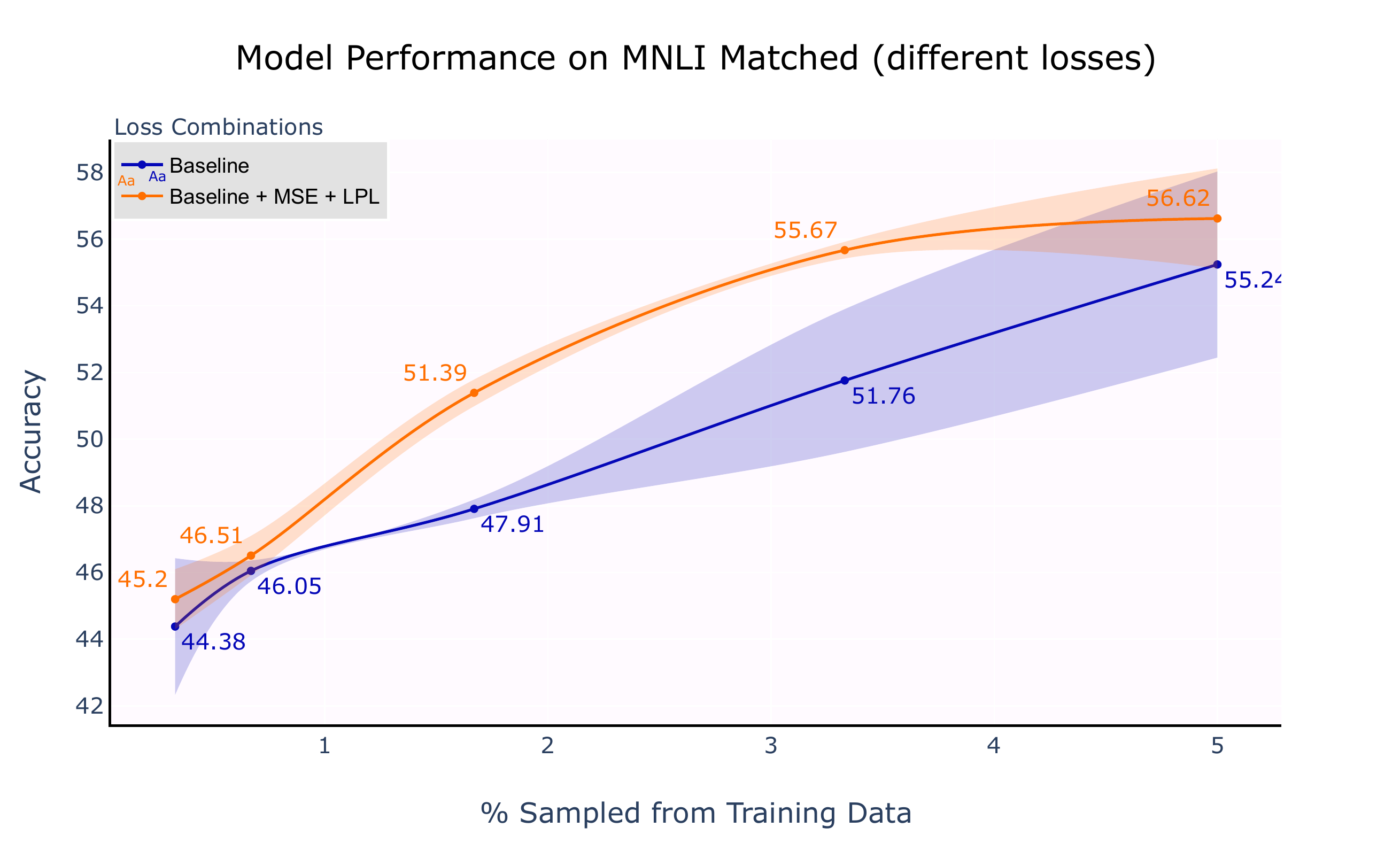}
\caption{Accuracy of alignment regularization on MNLI dataset with a varying number of \textit{matched} in-genre samples, up to $5\%$ of the training dataset only (total: $300$K samples).}
    \label{fig:exp_mnli_matched}
\end{figure}

\begin{figure}[t]
    \centering
    \includegraphics[width=\textwidth]{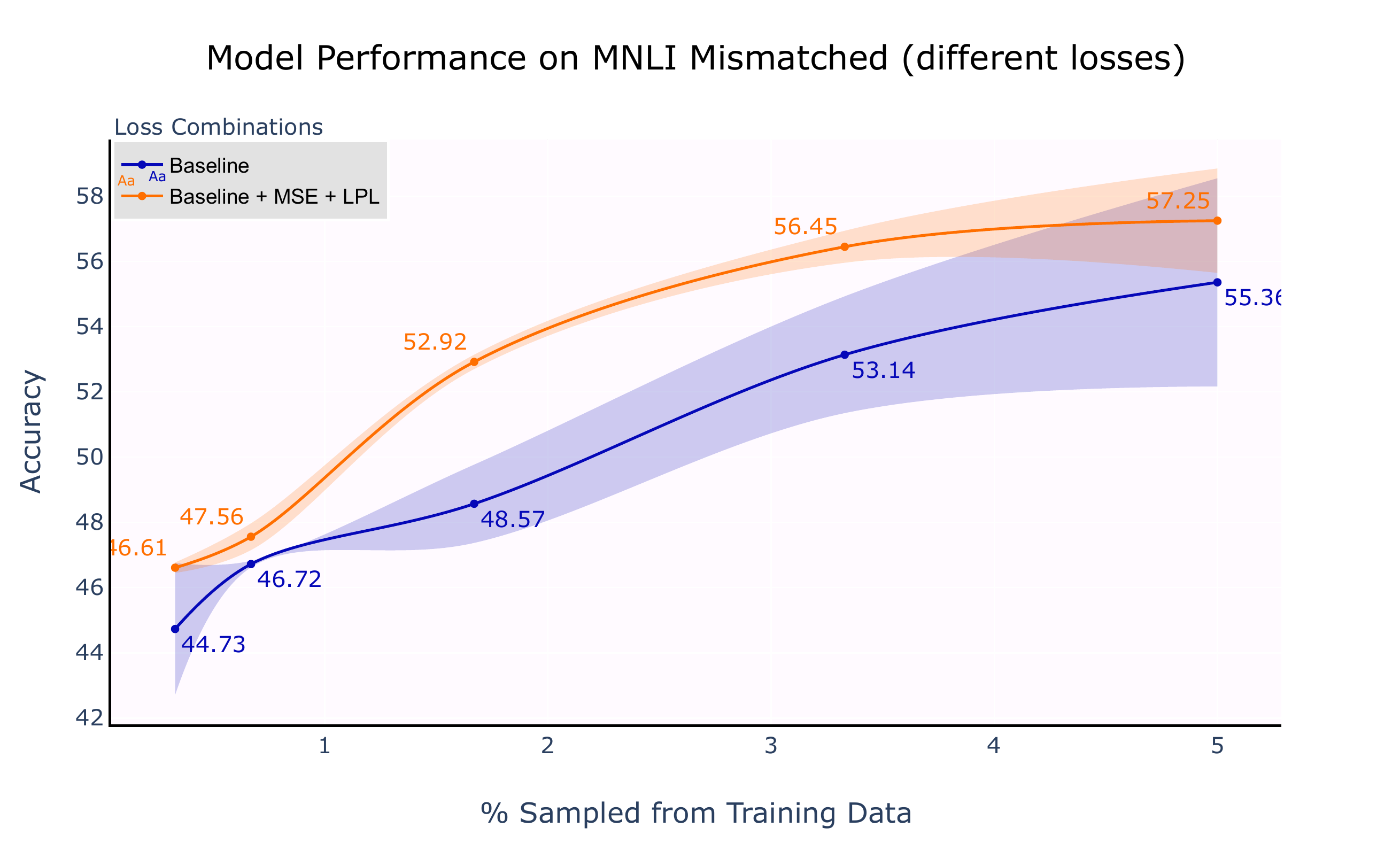}
\caption{Accuracy of alignment regularization on MNLI dataset with a varying number of \textit{mismatched} out-of-genre samples, up to $5\%$ of the training dataset only (total: $300$K samples).}
    \label{fig:exp_mnli_mismatched}
\end{figure}

To test the effectiveness of alignment as a regularizer, a $2$-layer FFN is used as shown in Figure \ref{fig:exp_align_network}; we measure the change in accuracy with respect to this baseline. An additional single layer network is utilized to perform the alignment with premise and hypothesis spaces. We experiment with the impact of the loss function on two datasets: the Stanford natural language inference (SNLI) \citep{bowman2015large} and the multigenre natural language inference dataset (MNLI) \citep{N18-1101}. SNLI consists of $500$K sentence pairs while MNLI contains about $433$k pairs. The MNLI dataset contains two test datasets. The \textit{matched} dataset contains sentences that are sampled from the same genres as the training samples while \textit{mismatched} samples test the models accuracy for out of genre text.


Figures \ref{fig:exp_snli_perf}, \ref{fig:exp_mnli_matched}, and \ref{fig:exp_mnli_mismatched} show the accuracy of the models when optimized with a standard cross-entropy loss (baseline) and with \texttt{MSE} and \texttt{LPL} combined. The accuracy is measured when the size of the training set is reduced.\footnote{\label{all_data_graph} Model accuracy using \texttt{MSE} and \texttt{MSE + LPL} with $100\%$ of the training data for STS-B is provided in Appendix A.1 and A.2 for SNLI / MNLI.} The reduced datasets are created by randomly sampling the required number from the entire dataset. The graphs show that an alignment loss consistently boosts accuracy of the model with respect to the baseline. The difference in accuracy (in Figure \ref{fig:exp_snli_perf}) is larger initially, it reduces as the training set becomes larger. This is because we calculate the neighbors for each premise from the training dataset only rather than any external text like Wikipedia (i.e., generate embeddings for Wikipedia sentences and then use them as neighbors). As the training size increases \texttt{LPL} has diminishing returns, as the neighbors tend to be part of the training pairs themselves.

\subsection{Crosslingual Word Alignment}
The cross lingual word alignment dataset is from \citet{dinu2014improving}. The dataset is extracted from the Europarl corpus and consists of word pairs split into training (5k pairs) and test (1.5k pairs) respectively.\footnote{http://opus.lingfil.uu.se/} From the 5K word pairs available for training only 3K pairs are used to train the model with \texttt{LPL} and an additional 150 pairs are used as the validation set (in case of Finnish 2.5K pairs are used). This is a reduced set in comparison to the models in  \Cref{tab:exp_alignment_performance} that are trained with all pairs.`

Compared to previous methods that look at explicit mapping of points between the two spaces, \texttt{LPL} tries to maintain the relations between words and their neighbors in the source domain while projecting them into the target domain. Along with the mapping methods in \Cref{tab:exp_alignment_performance}, previous methods also apply additional pre/post processing tranforms on the word embeddings as documented in \citet{artetxe2018generalizing} (described in  \Cref{tab:exp_alignment_steps_desc}). Cross-domain similarity local scaling (CSLS) \citep{conneau2017word} is used to retrieve the translated word.  \Cref{tab:exp_alignment_performance} shows the accuracy of our approach in comparison to other methods.

The accuracy of our proposed approach is better or comparable to previous methods that use similar numbers of transforms. It is similar to \citet{artetxe2018generalizing} while having fewer preprocessing steps. This is because we choose to optimize using gradient descent as compared to a matrix factorization approach. Thus, our implementation of \citet{artetxe2016learning} (\texttt{MSE} Loss only) under performs in comparison to the original baseline while giving improvements with \texttt{LPL}. Gradient descent has been adopted in this case because the loss function can be easily adopted by any neural network architecture in the future as compared to matrix factorization methods that will force architectures to use a two-step training process.

\subsection{Ablation Study}
\begin{figure}[t]
    \centering
    \includegraphics[width=\textwidth]{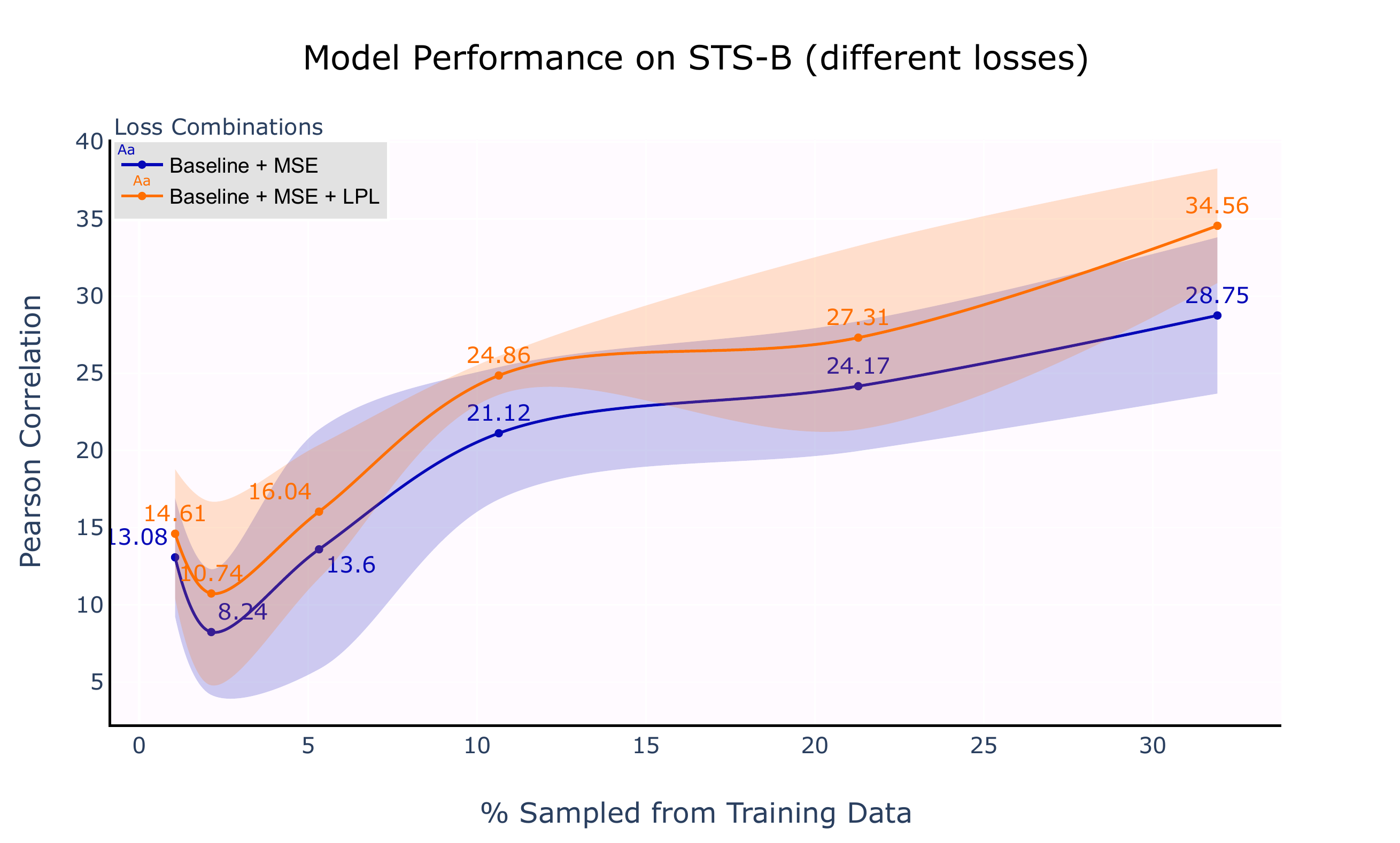}
\caption{Pearson correlation of alignment regularization on STS-B up to $50\%$ of the training dataset only (total: 4700 samples). The models are trained with \texttt{MSE} and \texttt{MSE + LPL}.}
    \label{fig:exp_stsb_mse_lpl_comparison}
\end{figure}

Although we introduce \texttt{LPL} in this paper, models in \cref{subsec:exp_sts} and \cref{subsec:exp_nli} are both trained with a combination of \texttt{MSE} and \texttt{LPL}. This raises the question: how much does \texttt{LPL} contribute to the overall performance of the model? We analyze this question by training a model separately on the STS-B dataset with \texttt{MSE} only and then comparing it with the prior model trained with the combined losses. In Figure \ref{fig:exp_stsb_mse_lpl_comparison}, we see that the model trained with \texttt{MSE + LPL} performs better with a maximum of $20.2\%$ relative improvement (sampled dataset size is $30\%$) over one trained with \texttt{MSE} alone. Additionally when the dataset size is small (less than $10\%$ of the training data), it is observed that variation in accuracy is smaller for the model with the combined loss. 

\begin{figure}[t]
    \centering
    \includegraphics[width=\textwidth]{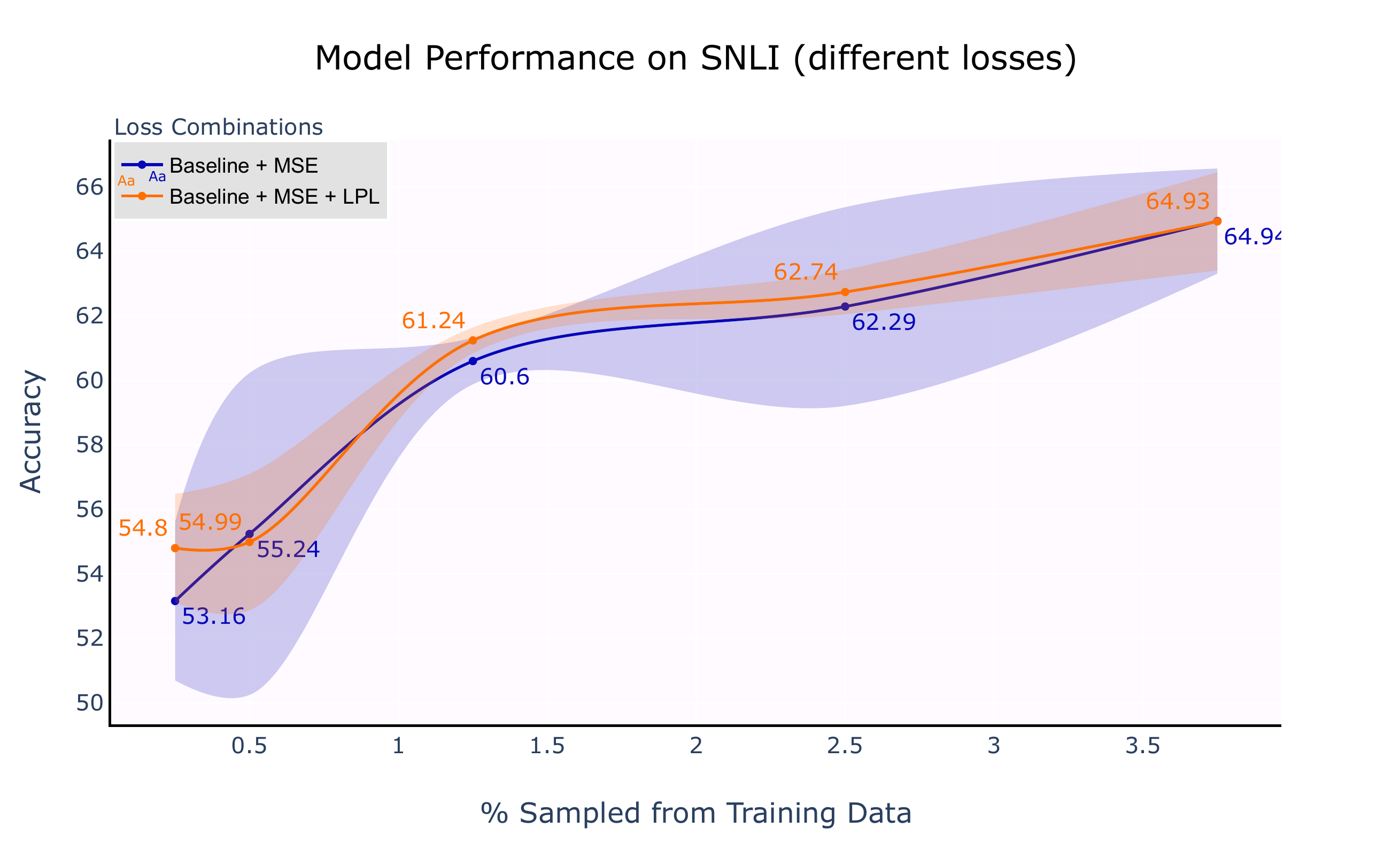}
\caption{Accuracy of alignment regularization on SNLI up to $5\%$ of the training dataset only. The models are trained with \texttt{MSE} and \texttt{MSE + LPL}.}
    \label{fig:exp_snli_mse_lpl_comparison}
\end{figure}

\begin{figure}[t]
    \centering
    \includegraphics[width=\textwidth]{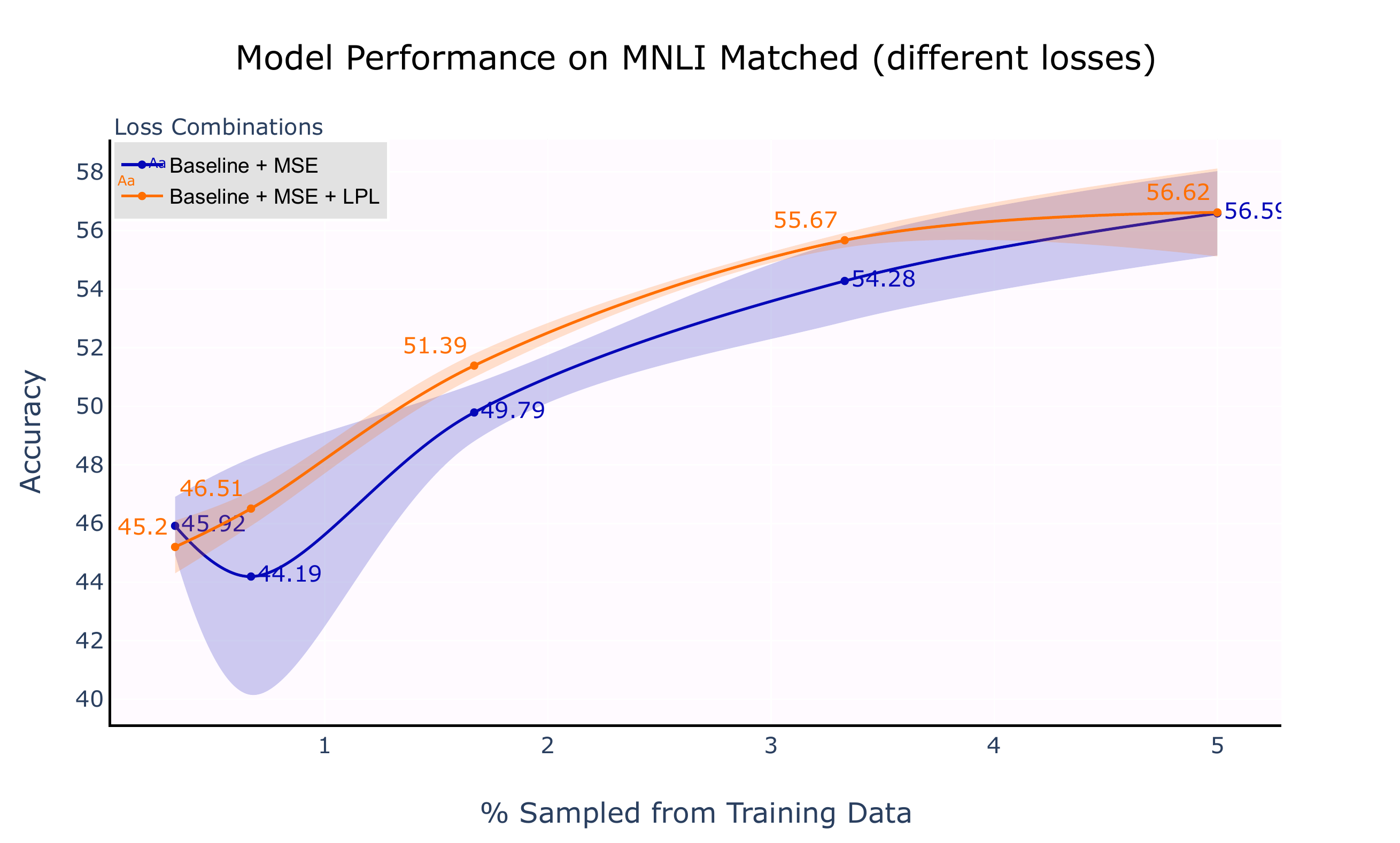}
\caption{Accuracy of alignment regularization on STS-B up to $5\%$ of the training dataset only. The models are trained with \texttt{MSE} and \texttt{MSE + LPL}.}
    \label{fig:exp_mnli_matched_mse_lpl_comparison}
\end{figure}

\begin{figure}[t]
    \centering
    \includegraphics[width=\textwidth]{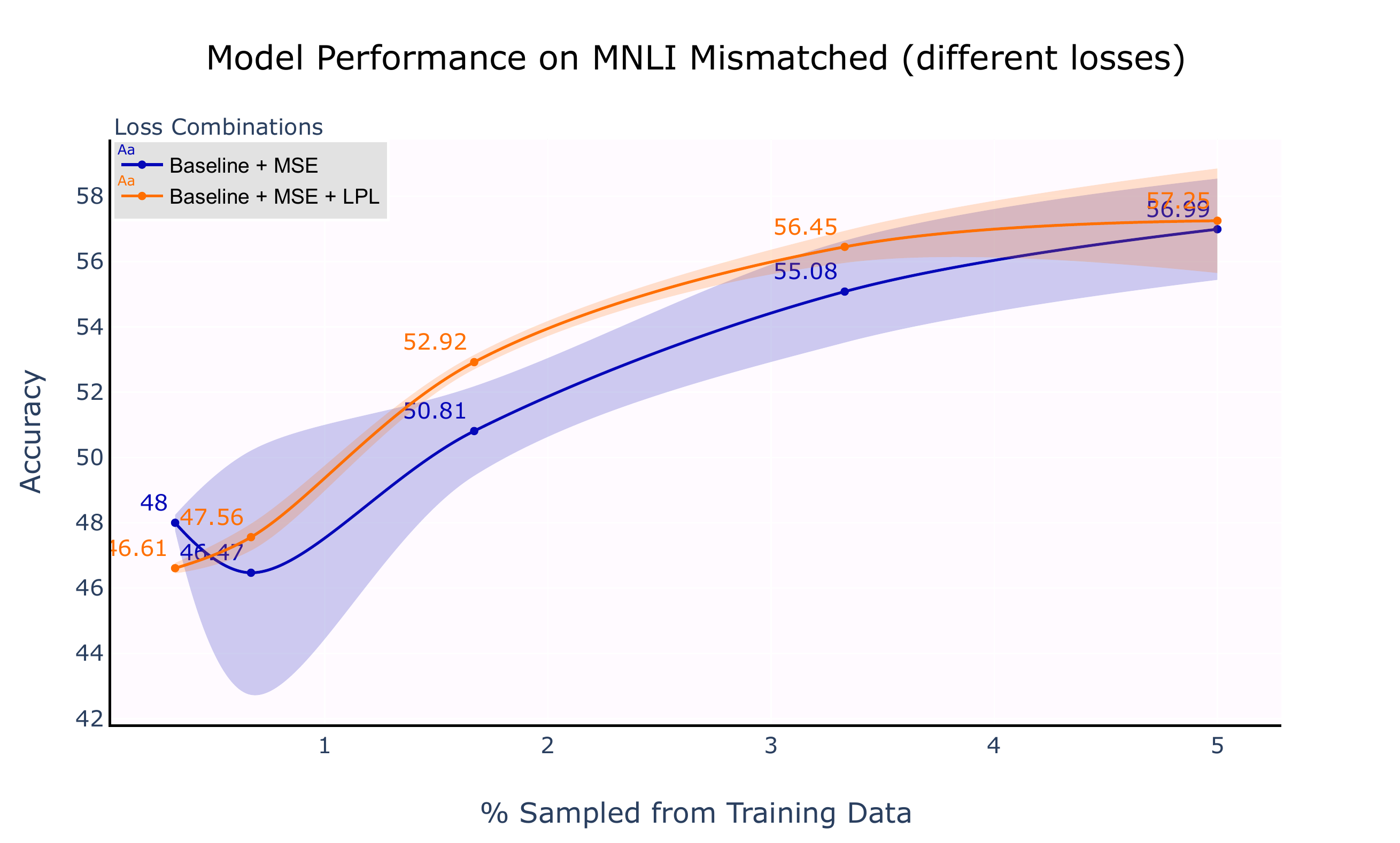}
\caption{Accuracy of alignment regularization on STS-B up to $5\%$ of the training dataset only. The models are trained with \texttt{MSE} and \texttt{MSE + LPL}.}
    \label{fig:exp_mnli_mismatched_mse_lpl_comparison}
\end{figure}

Additionally, we show the results of ablation studies on SNLI, MNLI (matched) and MNLI (mismatched). As seen in figures \ref{fig:exp_snli_mse_lpl_comparison}, \ref{fig:exp_mnli_matched_mse_lpl_comparison}, \ref{fig:exp_mnli_mismatched_mse_lpl_comparison}, \texttt{LPL}'s contribution to the model's accuracy in these tasks is lower in comparison to its contribution STS-B, but the variation in accuracy is smaller with it. Thus, we can conclude that \texttt{LPL} makes the model's performance consistent across experiments.

\subsection{Discussion}
\begin{table}  
    \centering
    \resizebox{\columnwidth}{!}{
    \begin{tabular}{c | p{.5cm} | l}
        \hline
        \textbf{Model Type} & \textbf{ID} & \textbf{Sentence}\Tstrut\\
        \hline
        & P & Family members standing outside a home.\Tstrut\\
        & H & A family is standing outside.\Tstrut\\
        \hline
        \multirow{4}{*}{B} & 1P & People standing outside of a building. \Tstrut\\
        & 1H & One person is sitting inside.\Tstrut\\
        & 2P & Airline workers standing under a plane.\Tstrut\\
        & 2H & People are standing under the plane.\Tstrut\\
        \hline
        \multirow{4}{*}{BM} & 3P & A group of four children dancing in a backyard.\Tstrut\\
        & 3H & A group of children are outside.\Tstrut\\
        & 4P & People standing outside of a building.\Tstrut\\
        & 4H & One person is sitting inside.\Tstrut\\
        \hline
        \multirow{4}{*}{BML} & 5P & A family doing a picnic in the park.\Tstrut\\
        & 5H & A family is eating outside.\Tstrut\\
        & 6P & Airline workers standing under a plane.\Tstrut\\
        & 6H & People are standing under the plane.\Tstrut\\
        \hline
    \end{tabular}
    }
    \caption{\textbf{Nearest neighbors extracted from SNLI classifier for a sentence pair representation.} P and H are the sample premise and hypothesis pair. The original label is \textit{Entailment}. (nP, nH) are the nearest neighbors of this sentence pair's representation from the penultimate layer of each classifier, i.e., baseline (B), baseline + \texttt{MSE} (BM), and baseline + \texttt{MSE + LPL} (BML). 1 \& 2 are nearest neighbors from the baseline, 3 \& 4 are when trained with \texttt{MSE} only and 5 \& 6 are when trained with \texttt{MSE} and LPL.}
    \label{tab:snli_nn_analysis}
\end{table}

\Cref{tab:snli_nn_analysis} shows the 2 nearest neighbors for a premise-hypothesis pair (P, H) taken from each classifier, i.e., baseline (B), baseline + \texttt{MSE} (BM), and baseline + \texttt{MSE + LPL} (BML) after they are trained (the dataset size is small at just 2000 samples).
Since NLI is a reasoning task, the sentence pair representations ideally will cluster around a pattern that represents \textit{Entailment} or \textit{Contradiction} or \textit{Neutral}. Instead what is observed is that when the samples are limited, sentence pair representations have NNs that are syntactically similar (NNs 1 and 2) for the baseline model. The predicted labels for the NN pairs are not clustered into entailment but are a combination of all 3. This problem is reduced for models trained with BM and BML (NNs 3 and 4 for BM, NNs 5 and 6 for BML). The predicted labels of the NNs are clustered into entailment only. The sentence pair representations cluster containing a single label suggest the models are better at extracting a pattern for entailment (and improving the model's ability to reason). This semantic clustering of representations can be attributed to the initial alignment (or divergence) between the premise and hypothesis. Also, we observe that a model regularized with \texttt{MSE} and \texttt{LPL} are more likely to reach optimal parameters consistently.\footnote{Check appendix section B for more details.}



\section{Conclusion}
In this paper, we introduce a new locality preserving loss (LPL) function that learns a linear relation between the given word and its neighbors and then utilizes it to learn a mapping for the neighborhood words that are not a part of the word pairs (parallel corpus). Also, we show how the results of the method are comparable to current supervised models while requiring a reduced set of word pairs to train on. The models are trained with SGD as compared to others that learn with SVD. Additionally, the same alignment loss is applied as a regularizer in a classification task (NLI) and a regression task (STS-B) to demonstrate how it can improve the accuracy of the model over the baseline.

\section*{Broader Impact}
In this section, we discuss the potential benefits and risks of using a locality preserving loss in the NLP tasks described in this paper.

\textbf{Benefits of LPL.} The main motivation of our work is to train models with limited data and showcase the effectiveness of locality preserving loss. When the dataset is small, models overfit the training data unable to generalize and exacerbate language biases. The main benefit of using \texttt{LPL} is that it maintains relationships between a datapoint and its neighbor in the target embedding space, restricting the model from overfitting.

\textbf{Risks of Utilizing LPL.} \texttt{LPL}'s ability to maintain relationships between points that are present in the source manifold after projection, can also be a risk. The weights $W$ in equation \ref{eq:alignment_lle} are learned by constructing a linear map between a datapoint and its neighbors using embeddings extracted from a pretrained model. Thus, any biased relationships prevalent in the pretrained model will become of a component of \texttt{LPL} and ultimately a part of the downstream fine-tuned model too. Hence it is necessary to evaluate the pretrained model thoroughly prior its use with \texttt{LPL}.

\section*{Acknowledgements}
We would like to thank our anonymous reviewers from the Adapt-NLP Workshop and previous NLP conferences for their constructive reviews. We thank Prof. Konstantinos Kalpakis for his insights about manifold alignment and locality preservation methods. The hardware used in our computational studies is part of the UMBC HPCF facility and UMBC's CARTA lab. This material is also based on research that is in part supported by the Air Force Research Laboratory (AFRL), DARPA, for the KAIROS program under agreement number FA8750-19-2-1003. The U.S.Government is authorized to reproduce and distribute reprints for Governmental purposes notwithstanding any copyright notation thereon. The views and conclusions contained herein are those of the authors and should not be interpreted as necessarily representing the official policies or endorsements, either express or implied, of the Air Force Research Laboratory (AFRL), DARPA, or the U.S. Government.


\bibliography{references/manifold_alignment.bib,references/word_embeddings.bib}
\bibliographystyle{acl_natbib}

\newpage
\clearpage

\appendix
\section*{Appendix}
\section{Experimentation Details}
\label{sec:appendix_a}
This section discusses in detail the various experiments conducted in this paper. For each set of experiments, we provide an overview of the dataset used, a description of the dataset (with samples), information about the training set and list of any hyper-parameters that are optimized. The computing infrastructure used is a single Tesla P100-SXM2 GPU to train a single model. The GPU consumption is driven by the base text / word encoding model. GPU usage while training the feedforward network (FFN) is limited. In practice, once the embeddings are extracted from an language encoder (like BERT), multiple FFs can be simultaneously trained on a single GPU. Additional GPUs are used to scale experiments while training models with different loss function combinations and dataset sizes.

\subsection{Semantic Text Similarity (STS)}
In order to understand the impact of Locality Preserving Loss (LPL) while finetuning a model to measure how similar two sentences are, we use the STS-B dataset \citep{cer2017semeval} that is widely utilized as a part of the GLUE benchmark \citep{wang2018glue}. STS-B consists of a total of $8628$ pairs of sentences of which $5749$ are for training, $1500$ pairs are part of the dev set (for hyper-parameter tuning) and $1379$ are test pairs. The labels are a continuous value between $0 - 5$ where $0$ represents sentences that are not similar while $5$ represents sentences are that have the same syntax and meaning. While training various models, the labels are normalized between $0$ and $1$.  \Cref{tab:app_sts_sample} shows a few examples of sentence pairs in the dataset.

\begin{table}[ht!]
    \centering
    \begin{tabular}{p{2.6cm} | p{2.6cm} | p{1cm}}
        \hline
         \textbf{Sentence 1} & \textbf{Sentence 2} & \textbf{Label}\Tstrut\\ 
        \hline
        A plane is taking off.  & An air plane is taking off. & 5.0 \Tstrut\\
        A man is spreading shreded cheese on a pizza.& A man is spreading shredded cheese on an uncooked pizza.& 3.8\\
        Three men are playing chess.& Two men are playing chess. & 2.6\\
        \hline
    \end{tabular}
  \caption{Sample sentence pairs from STS-B \citep{cer2017semeval} dataset with their corresponding labels. The labels represent subjective human judgements of how similar the sentences are. It is a continuous variable.}
  \label{tab:app_sts_sample}
\end{table}
A BERT model \citep{devlin2018bert} with an additional $2$-layer feedforward network (FFN) ($P_n$) predicts the similarity score between sentences and is optimized with MSE (between predicted score and label). The FFN consists of $2$ hidden layers, each of size $1024$. The baseline model (trained only with MSE) has a concatenated input (as shown in Figure \ref{fig:exp_align_network} - \textit{baseline}). While using an additional alignment loss (MSE or LPL), an additional single hidden layer FFN ($A_n$) is attached that aligns the two sentence manifolds. The aligned projection of sentence $1$ is then concatenated as input to $P_n$ (as shown in Figure \ref{fig:exp_align_network} - \textit{baseline with alignment}). Although this increases the number of parameters in the $P_n$ (by $768$ x $1024$) for models trained with alignment, we experimented with increasing the baseline model with the same number of hidden layer parameters and found the baseline's performance to decrease or remain constant. Hence, the size of the layers is maintained at $1024$. 

\subsubsection{Crossvalidation}
\label{subsubsec:app_sts_crossval}
Figure \ref{fig:exp_stsb_pearson} shows the pearson correlation between the predicted and actual similarity score. For each sample size on the X-axis, 3-fold cross validation is performed. Each time cross-validation is performed, different training pairs are randomly sampled (without replacement) from the complete dataset and the seed for initializing weights of each layer in the FFN is changed. To maintain consistency across experiments, the seed is maintained constant across trained models.

\subsubsection{Hyperparameters}
As described in \cref{subsec:align_as_reg}, the hyperparameters include $\gamma$ and $\delta$. $\gamma$ is a learning rate that defines how much of the alignment loss functions contribute to the overall loss. In practice, $\gamma$ is set to $1.0$. 

Equation \ref{eq:alignment_regularizer} uses LPL as a regularizer. We implement this as a contrastive divergence loss. 


\begin{equation}
\mathcal{L}_{\textrm{mcd}}^i = \gamma \sum_i{\max(-1.0, \delta_{y_i} * \mathcal{L}_{\textrm{mse}}^{i}})
\label{eq:alignment_mse_cd_loss}
\end{equation}

\begin{equation}
\mathcal{L}_{\textrm{lcd}}^i = \gamma \sum_i{\max(-1.0, \delta_{y_i} * \mathcal{L}_{\textrm{lpl}}^{i}})
\label{eq:alignment_lpp_cd_loss}
\end{equation}

The above equations change $\mathcal{L}_{\textrm{mse}}$ and $\mathcal{L}_{\textrm{lpp}}$ to a contrastive divergence (max margin) loss. We set the maximum margin to $0.1$. The margin is manually tuned after optimizing it on validation pairs. The overall learning rate to train the model is $0.0001$. The optimizer is RMSProp. 

$\delta$ is the label specific hyper-parameter that defines how much loss can be contributed by a specific label. As the STS label is a continuous variable, $\delta$ is equal to the label value. This forces sentence pairs with scores that are $5.0$ to have maximum loss while pairs with scores that tend towards $0$ create a negative squared loss (this moves the embeddings apart) limited up-to the maximum margin.

\subsubsection{Performance on Larger Dataset}
\label{subsec:app_stsb_large_dataset}
\begin{figure}[t]
    \centering
    \includegraphics[width=\textwidth]{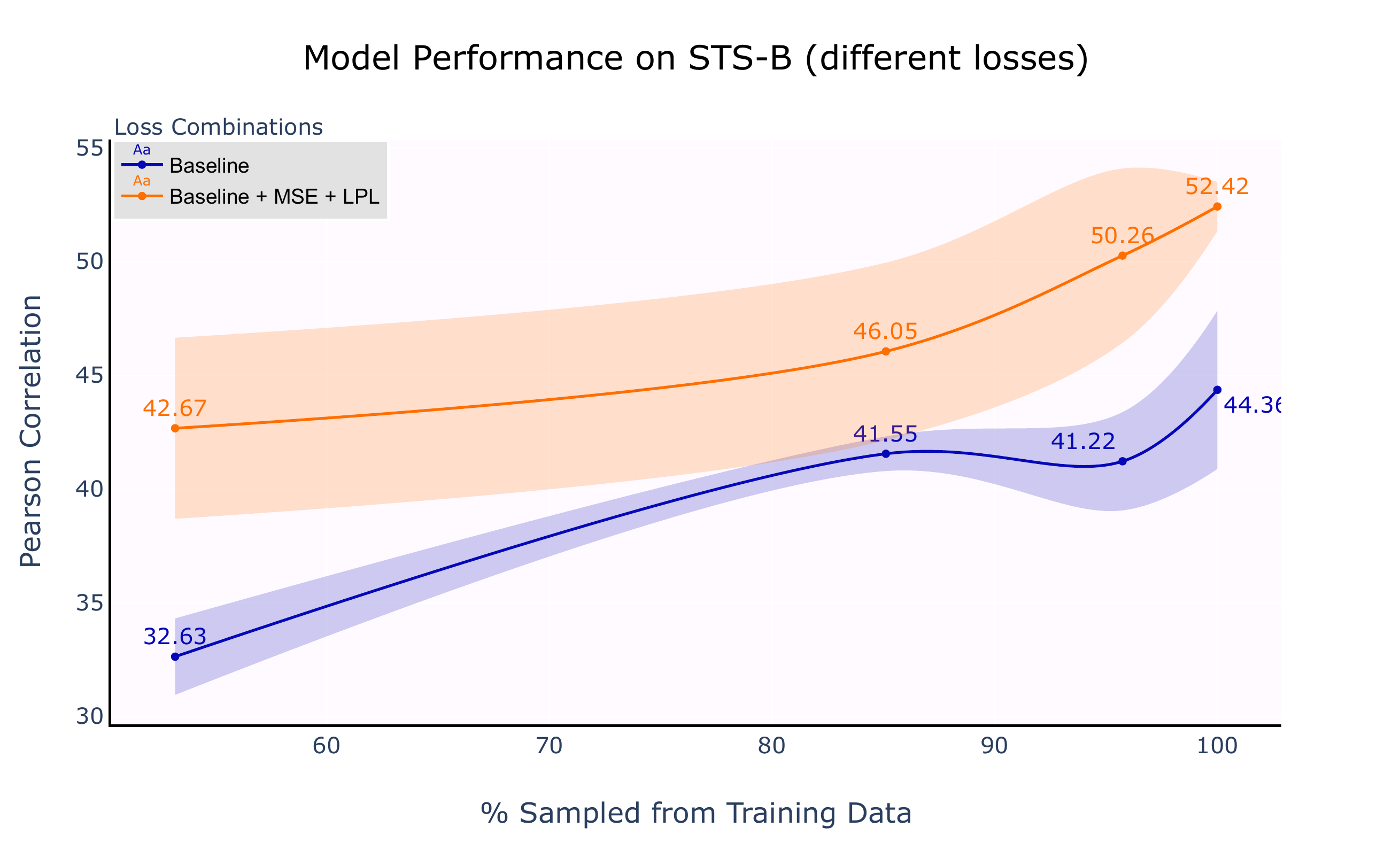}
    \caption{Pearson correlation on semantic text similarity with STS-B dataset for training sample size greater than $50\%$ of the original dataset.}
    \label{fig:exp_stsb_pearson_large}
\end{figure}
In Figure \ref{fig:exp_stsb_pearson_large}, we see that the \texttt{LPL + MSE} loss model performs better than the baseline when dataset size is increased. The lowest bound in performance is higher than the upper bound of the baseline. \texttt{LPL}s improvement over the baseline when the entire dataset is used shows that as a whole STS-B may benefit from \texttt{LPL} irrespective of the dataset size. This is because \texttt{LPL} explicitly models a relationship between sentences present in a given sample text's neighborhood, ensuring that those relationships are maintained while computing the similarity between a given sentence pair. 

\subsection{Natural Language Inference (NLI)}
\cref{subsec:exp_nli} discusses in detail the experiments with natural language inference. Experiments are conducted on SNLI \citep{snli:emnlp2015} and MNLI \citep{williams2017broad} datasets. Stanford's natural language inference dataset contains sentence pairs consisting of a premise and hypothesis. The model predicts if the hypothesis \textit{entails} or \textit{contradicts} the premise or if they have a neutral relationship (i.e., the two sentences are not related). The dataset contains $500K$ training pairs, $10K$ pairs in the dev set and $10K$ pairs of sentences for testing.  \Cref{tab:app_snli_sample} shows a few sentence pairs from the dataset.

\begin{table}[ht!]
    \centering
    \begin{tabular}{p{2.4cm} | p{2.4cm} | p{1.55cm}}
        \hline
         \textbf{Premise} & \textbf{Hypothesis} & \textbf{Label}\Tstrut\\ 
        \hline
        A person on a horse jumps over a broken down airplane.  & A person is at a diner, ordering an omelette. & \textit{Contradiction} \Tstrut\\
        \hline
        A person on a horse jumps over a broken down airplane. & A person is outdoors, on a horse. & \textit{Entailment}\\
        \hline
        Children smiling and waving at camera & They are smiling at their parents & \textit{Neutral}\\
        \hline
    \end{tabular}
  \caption{Samples from the SNLI \citep{snli:emnlp2015} dataset. Each pair consists of two sentences and a label with one of three values \textit{entailment}, \textit{neutral}, \textit{contradiction}.}
  \label{tab:app_snli_sample}
\end{table}

MNLI \citep{williams2017broad} is an extension of the SNLI dataset where the sentences are from multiple genres such as letters promoting non-profit organizations, government reports and documents as well as fictional books. This expands the variation in language used in sentences, reducing the model's ability to memorize sentence pairs and their labels.  \Cref{tab:app_mnli_sample} contains example pairs from the MNLI training data.

\begin{table}[ht!]
    \centering
    \begin{tabular}{p{2.7cm} | p{2.4cm} | p{1.2cm}}
        \hline
         \textbf{Premise} & \textbf{Hypothesis} & \textbf{Label}\Tstrut\\ 
        \hline
        Conceptually cream skimming has two basic dimensions - product and geography.&  Product and geography are what make cream skimming work.& \textit{Neutral}\\
        \hline
        How do you know? All this is their information again.& This information belongs to them.& \textit{Entailment}\\
        \hline
    \end{tabular}
  \caption{Samples from the SNLI \citep{snli:emnlp2015} dataset. Each pair consists of two sentences and label with one of three values \textit{entailment}, \textit{neutral}, \textit{contradiction}}
  \label{tab:app_mnli_sample}
\end{table}

The model's architecture is described in Figure \ref{fig:exp_align_network}. The sentence embeddings are extracted from BERT \citep{devlin2018bert} and the NLI classification layers are trained separately. For experiments in \cref{subsec:exp_nli}, the trained model is a $2$-layer FNN, each hidden layer with a size of $4096$. The activation function is a Leaky RELU. Similar to the experiments in \cref{subsubsec:app_sts_crossval}, we perform 3-fold crossvalidation where the training dataset is resampled and the results presented in \cref{subsec:exp_nli} are an average over these $3$ runs.

\subsubsection{Hyperparameters}
The models are trained with a learning rate of $0.0001$ and the optimizer is RMSProp. In order to train the model with LPL and MSE, $\delta$ is configured. In comparison to a continuous $\delta$ variable used in the STS task, in NLI, the $\delta$ is constant for each label. For each dataset, the $\delta$ parameter is set after manual testing on the dev set. Sentence pairs that have a \textit{neutral} label tend to be sentences that are dissimilar. The semantic difference can be based on differing subjects, predicates or objects in the sentence, the content / topic or even genre of the sentence. Hence while training the model, these projections are \textbf{separated} with the alignment loss (and have a negative $\delta$) rather than converged. 

From our initial experiments, we found that lower $\delta$ on \textit{entailment} and \textit{contradiction} yielded no change in accuracy from baseline. The accuracy increases when a higher positive $\delta$ scalar multiple is attached to \textit{entailment} \& \textit{contradiction}, and a negative scalar is multiplied to the loss generated for \textit{neutral} label samples.

\begin{table}[ht!]
    \centering
    \begin{tabular}{p{1.4cm} | p{1.4cm} | p{1.4cm} | p{1.7cm}}
        \hline
         \textbf{Dataset} & \textbf{Entailment} & \textbf{Neutral}\Tstrut & \textbf{Contradiction}\\ 
        \hline
        SNLI & $100.0$ & $-5.0$ & $0.0$\Tstrut\\
        MNLI & $250.0$ & $1.0$ & $-10.0$\\
        \hline
    \end{tabular}
  \caption{Samples from the SNLI \citep{snli:emnlp2015} dataset. Each pair consists of two sentences and label with one of three values \textit{entailment}, \textit{neutral}, \textit{contradiction}}
  \label{tab:app_delta_values}
\end{table}

\subsubsection{Performance on Larger Dataset}
\label{subsec:app_nli_large_dataset}
\begin{figure}[t]
    \centering
    \includegraphics[width=\textwidth]{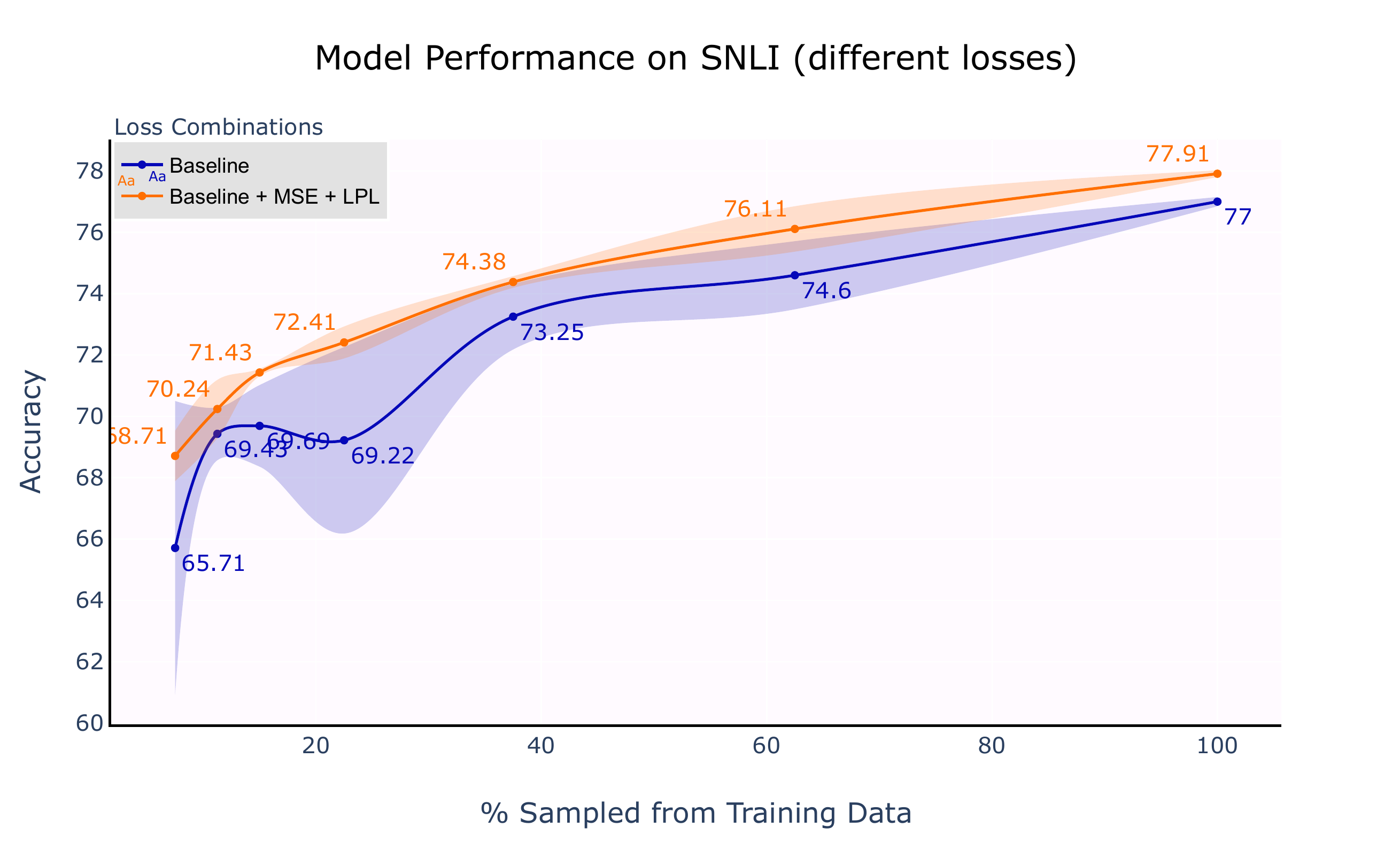}
\caption{Accuracy of alignment regularization on SNLI. The graph shows the accuracy, averaged across 3 runs, for differing size of training samples from $5\%$ to $100\%$ of the training dataset only (total: $500$K).}
    \label{fig:exp_snli_perf_large}
\end{figure}

\begin{figure}[t]
    \centering
    \includegraphics[width=\textwidth]{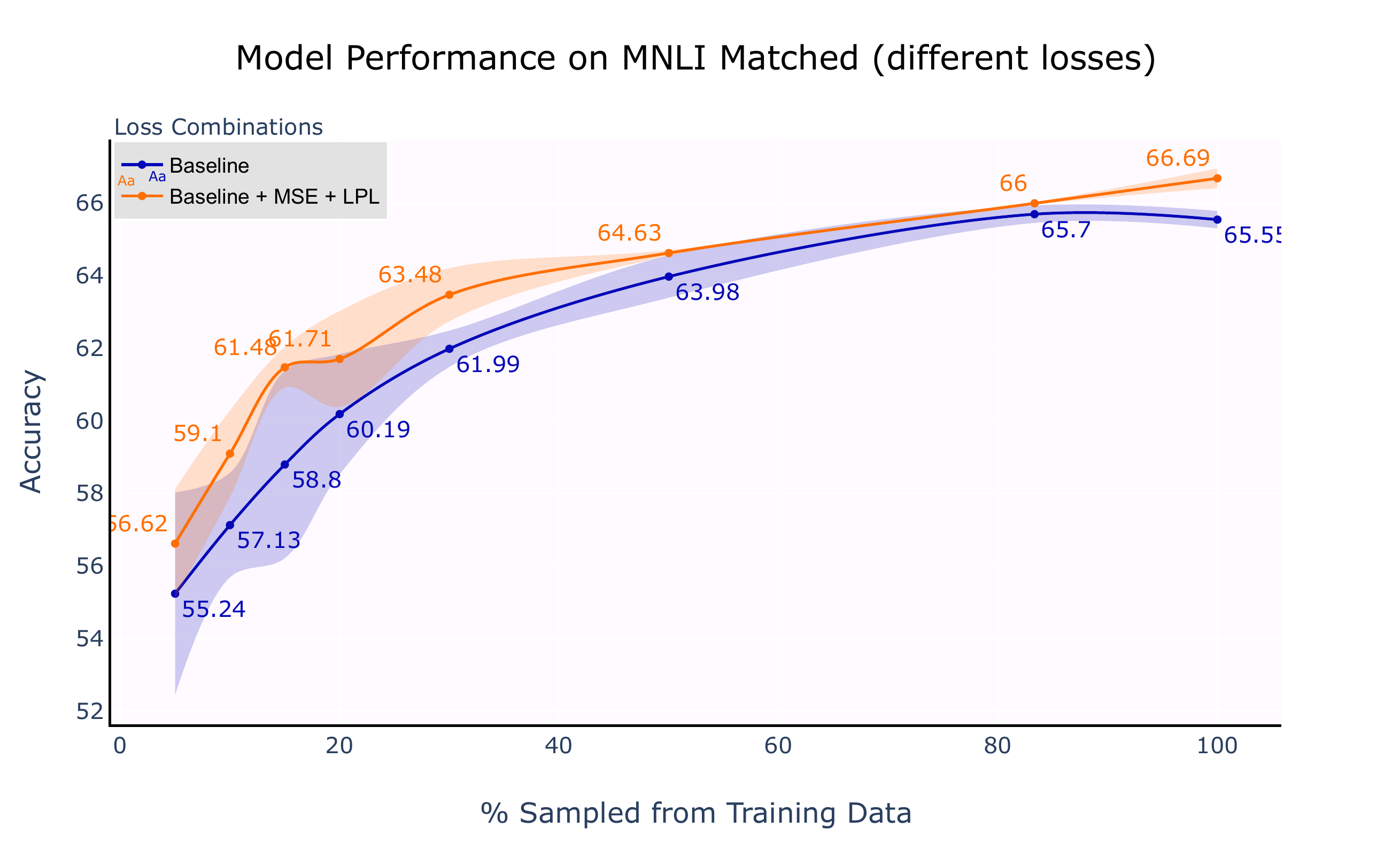}
\caption{Accuracy of alignment regularization on MNLI dataset with a varying number of \textit{matched} in-genre samples, from $5\%$ to $100\%$ of the training dataset only (total: $300$K samples).}
    \label{fig:exp_mnli_matched_large}
\end{figure}

\begin{figure}[t]
    \centering
    \includegraphics[width=\textwidth]{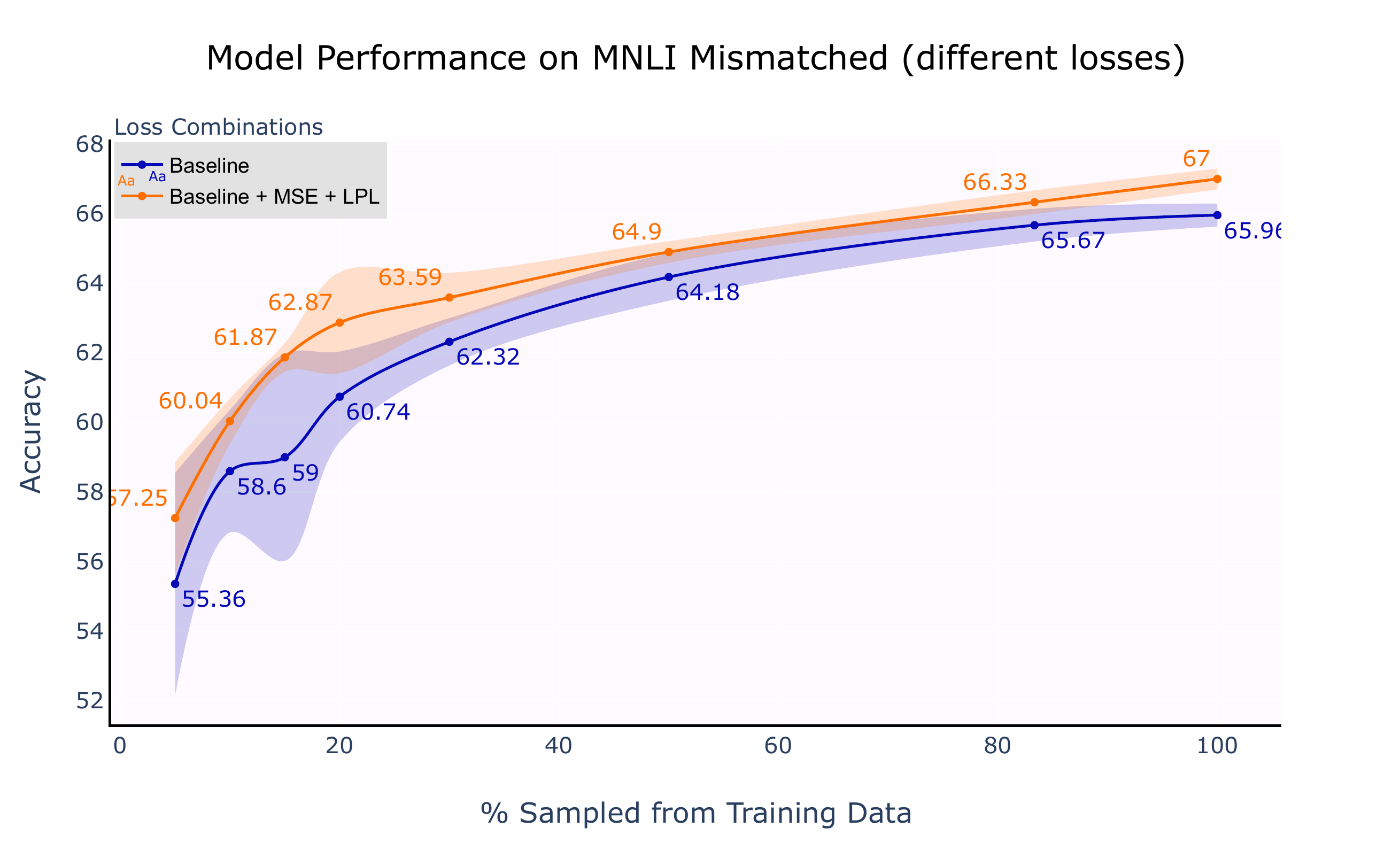}
\caption{Accuracy of alignment regularization on MNLI dataset with a varying number of \textit{mismatched} out-of-genre samples, from $5\%$ to $100\%$ of the training dataset only (total: $300$K samples).}
    \label{fig:exp_mnli_mismatched_large}
\end{figure}

Apart from better accuracy when the training dataset is small, in Figures \ref{fig:exp_snli_perf_large},\ref{fig:exp_mnli_matched_large}, and \ref{fig:exp_mnli_mismatched_large}, we observe the accuracy of models trained with the alignment loss using \texttt{MSE} alone and another in combination with \texttt{LPL} converge as the number of training samples increases. This happens because of the way k-nearest neighbor (k-NN) is computed for each embedding in the source domain. We use BERT to generate the embedding of each sentence in the SNLI and MNLI dataset. Because BERT itself is trained on millions of sentences from Wikipedia and Book Corpus, searching for k-NN embeddings for each sentence from this dataset (for each sentence in the training sample) is computationally difficult. In order to make the k-NN search tractable, neighbors are extracted from the dataset itself (500K sentences in SNLI and 300K sentences in MNLI). This impacts the overall improvement in accuracy using \texttt{LPL} as it is not a perfect reconstruction of the datapoint (using its neighbors). Initially when the dataset is small the neighbors are unique. As the dataset size increases, the unique neighbors reduce and are subsumed by the overall supervised dataset (hence \texttt{MSE} begins to perform better). The impact of \texttt{LPL} reduces as the number of unique neighbors decreases and the entire dataset is used to train the model. This is unlikely to happen when NNs from a larger unrelated text corpus reconstruct local manifolds.

\begin{figure*}
    \centering
    \includegraphics[width=\textwidth]{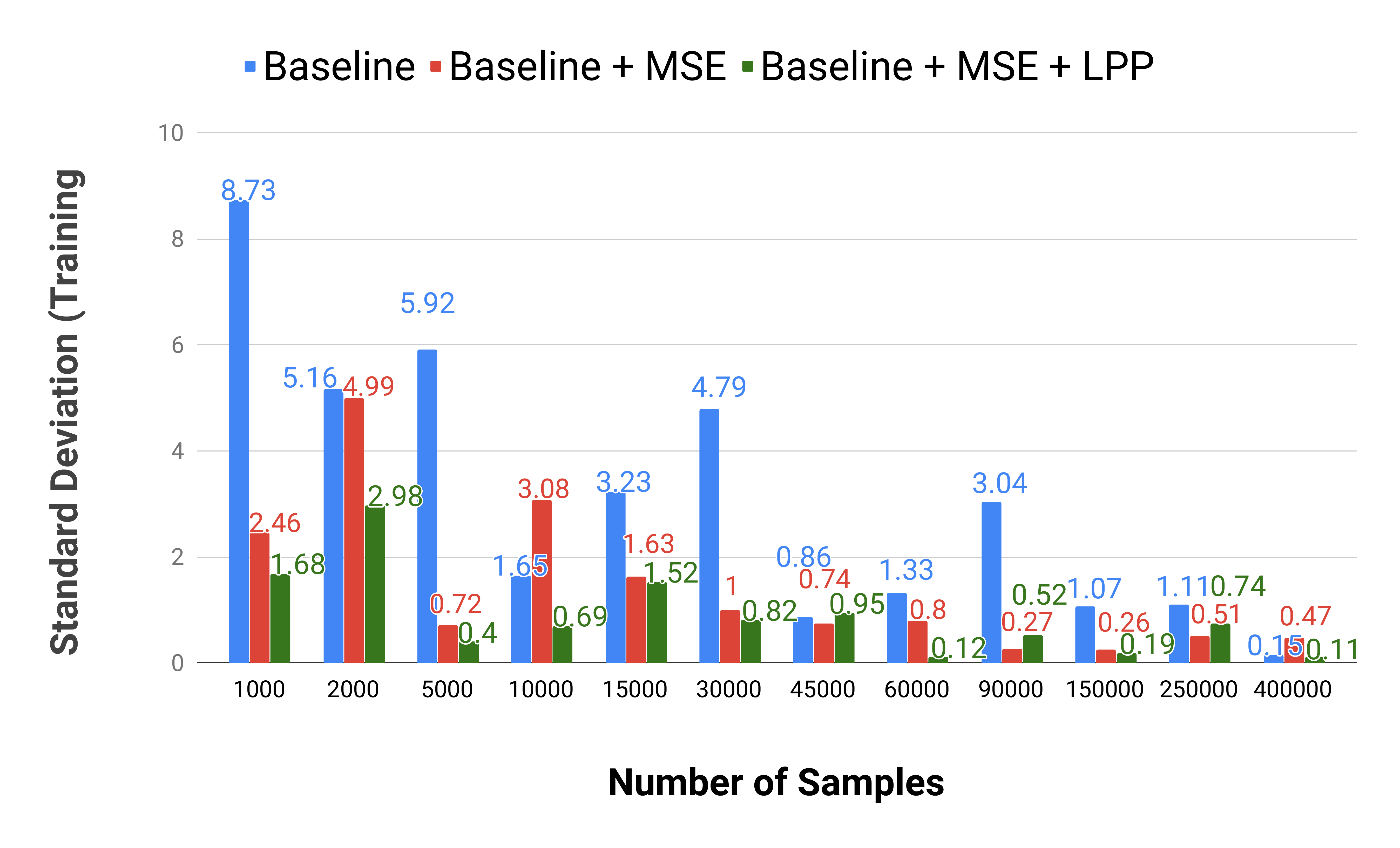}
\caption{\textbf{Standard deviation in performance}. The graphs show the standard deviation in performance over 3 runs when the SNLI training dataset size varies In each case, the model trained with \texttt{LPL + MSE} + task loss has the least variation in performance, while the model trained with task loss + \texttt{MSE} or the task loss alone having a higher variance in performance.}
    \label{fig:exp_snli_stdev}
\end{figure*}

\begin{figure*}
    \centering
    \includegraphics[width=\textwidth]{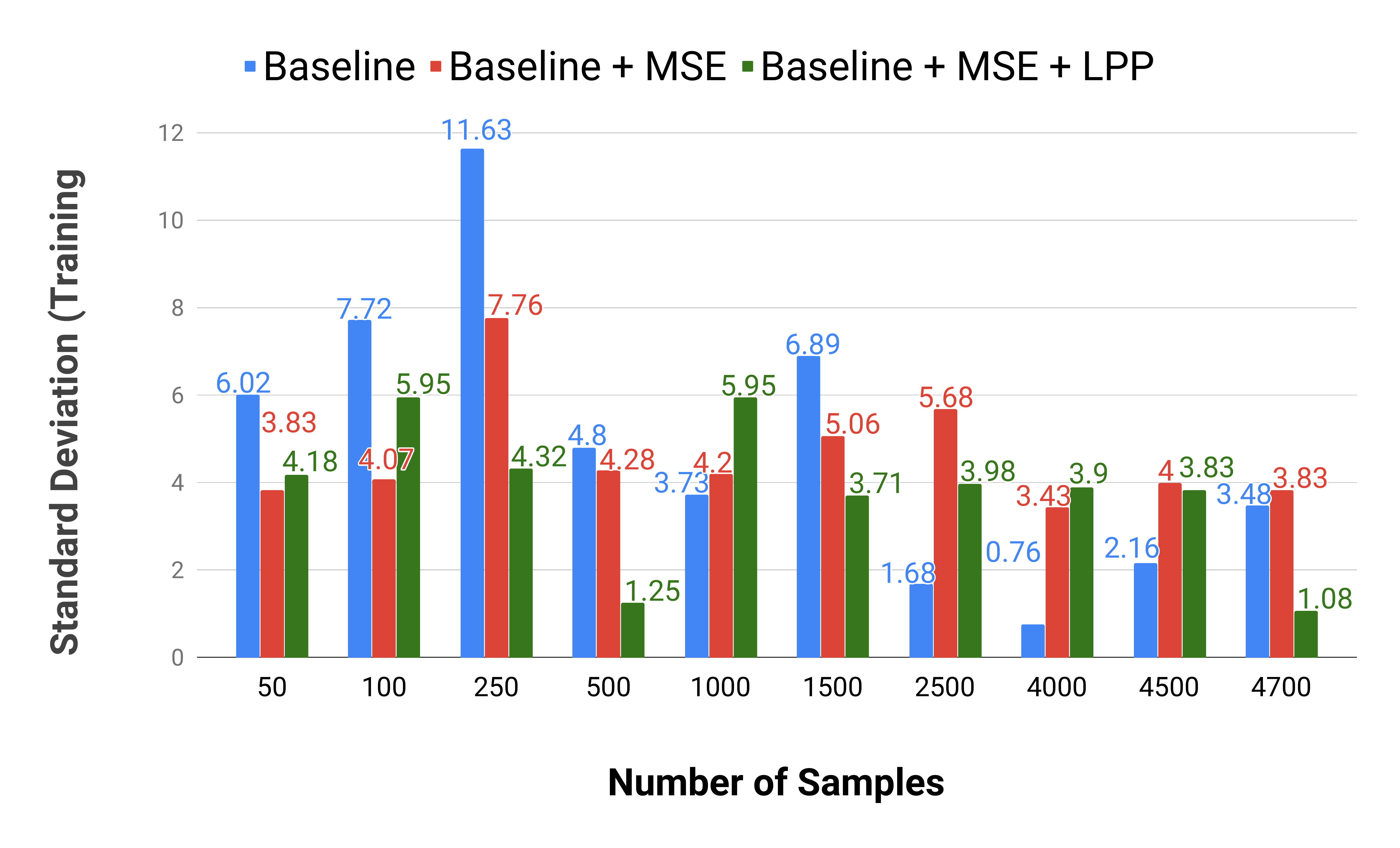}
\caption{\textbf{Standard deviation in pearson correlation}. The graphs show the standard deviation in correlation over 3 runs when the STS-B training dataset size varies. In each case, the model trained with LPL + MSE + task loss has the least variation in performance, while the model trained with task loss + MSE or the task loss alone having a higher variance in performance.}
    \label{fig:exp_stsb_stdev}
\end{figure*}

\subsection{Crosslingual Word Alignment}
\begin{figure*}
    \centering
    \includegraphics[scale=0.52]{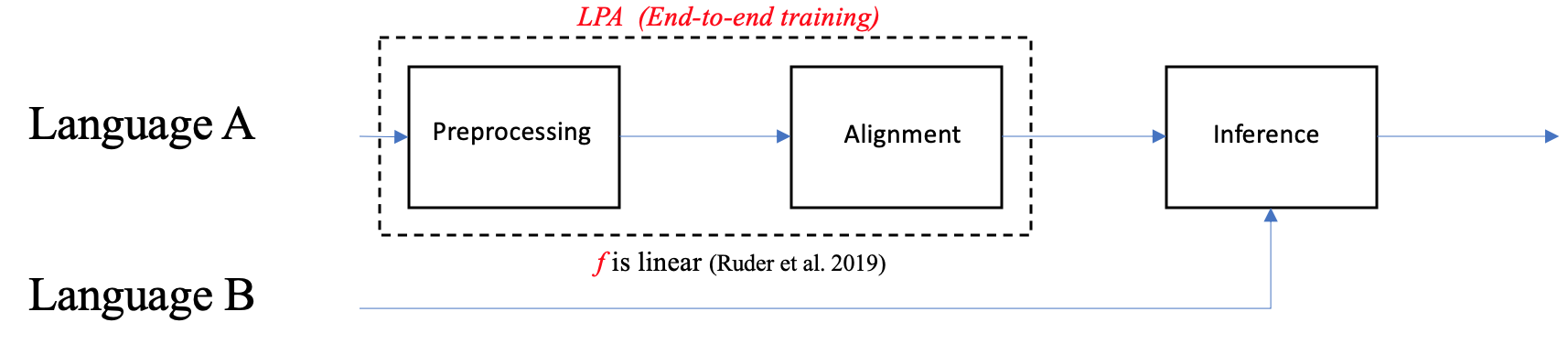}
\caption{\textbf{Crosslingual Word Alignment Pipeline (CLA).} As described in \citet{ruder2019survey}, the CLA pipeline includes choosing an existing set of pretrained vectors that can be preprocessed.  \Cref{tab:exp_alignment_steps_desc} contains a list of preprocessing functions that can be applied. \citet{artetxe2018generalizing} evaluate each preprocessing method in detail. Alignment involves learning the transform matrix $W$ and inference involves finding the translated word.}
    \label{fig:app_cla_pipeline}
\end{figure*}

For experiments with cross lingual word alignment, we use the parallel corpus from \citet{dinu2014improving}.  \Cref{tab:app_acc_en_it_comparison} compares our method with other methods such as \citet{xing2015normalized} and \citet{faruqui2014improving} and extends results in  \Cref{tab:exp_alignment_performance}. The dataset contains word pairs collected from the Europarl corpus. The bilingual dictionary has 5000 word pairs for training and 1500 word pairs for testing and evaluation. The language pairs in the dictionary include English $\leftrightarrow$ Italian, English $\leftrightarrow$ German, English $\leftrightarrow$ Finnish and English $\leftrightarrow$ Spanish. Figure \ref{fig:app_cla_pipeline} shows the pipeline to perform cross-lingual word alignment.

\begin{table}[h]
    \centering
    \begin{tabular}{p{5cm} | p{1cm}}
        \hline
         \textbf{Method} & \textbf{EN - IT}\Tstrut\\ 
        \hline
        \citet{mikolov2013distributed} & $34.93$\Tstrut\\
        \citet{xing2015normalized} &  $36.87$\\
        \citet{faruqui2014improving} &  $37.80$\\
        \citet{artetxe2016learning} & $39.27$\Tstrut\\
        \hline
        Locality Preserving Loss (Our Work) & \textbf{43.33}\Tstrut\\
        \hline
    \end{tabular}
  \caption{Accuracy of various models predicting word translated from English to Italian.}
  \label{tab:app_acc_en_it_comparison}
\end{table}

Once the initial pretrained word vectors are selected, preprocessing can be applied. Preprocessing functions include unit normalization, whitening and z-normalization ( \Cref{tab:exp_alignment_steps_desc}). \citet{ruder2019survey} observe that deep neural networks are perform word alignment poorly in comparison to a linear tranformation ($Y = WX$). The linear transformation matrix $W$ is learned by minimizing the sum of squared loss between $Y$ and $WX$. Optimal parameters for $W$ are learned by minimizing the loss with singular value decomposition (SVD) \citep{artetxe2016learning}. 

Locality Preserving Alignment combines preprocessing functions and training $W$. Also, parameters of $W$ are learned using SGD, showcasing that LPL can be added to both linear and non-linear transformations. To find the translated word in the target embedding space, multiple inference mechanism are available such as nearest neighbor (NN), inverted softmax \citep{smith2017offline}, and cross-domain similarity scaling (CSLS) \citep{conneau2017word}. We use \textbf{CSLS} to find the translated word in the target language. 

In our experiments, we use original pretrained word embeddings obtained from the dataset \citep{dinu2014improving}. The models are trained with a learning rate of $0.001$. The $\beta$ parameter is the learning rate specific to LPL (from equation \ref{eq:alignment_total_loss_ph2}) and is set to $0.7$ and is manually tuned against the validation dataset.

\begin{figure*}
    \centering
    \includegraphics[width=\textwidth]{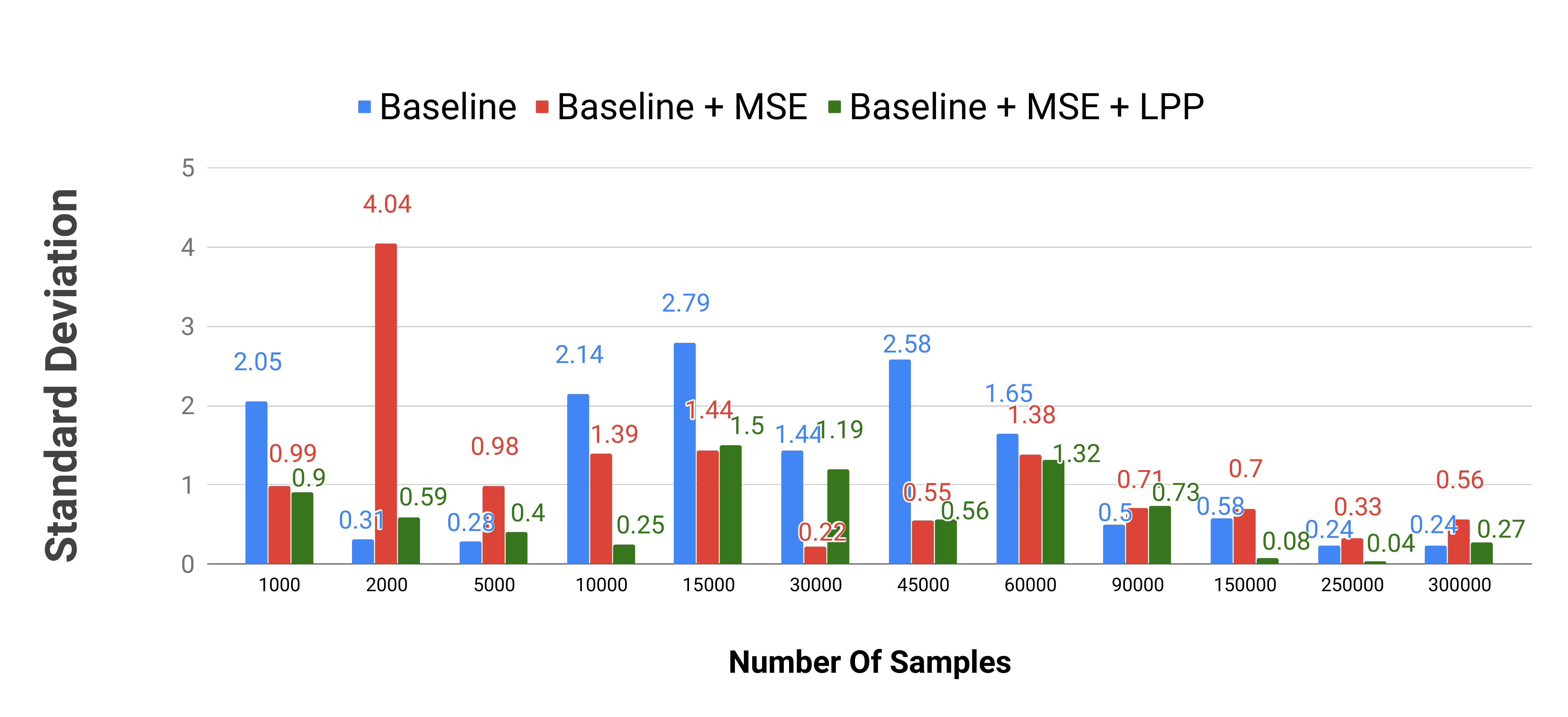}
    ~
    \includegraphics[width=\textwidth]{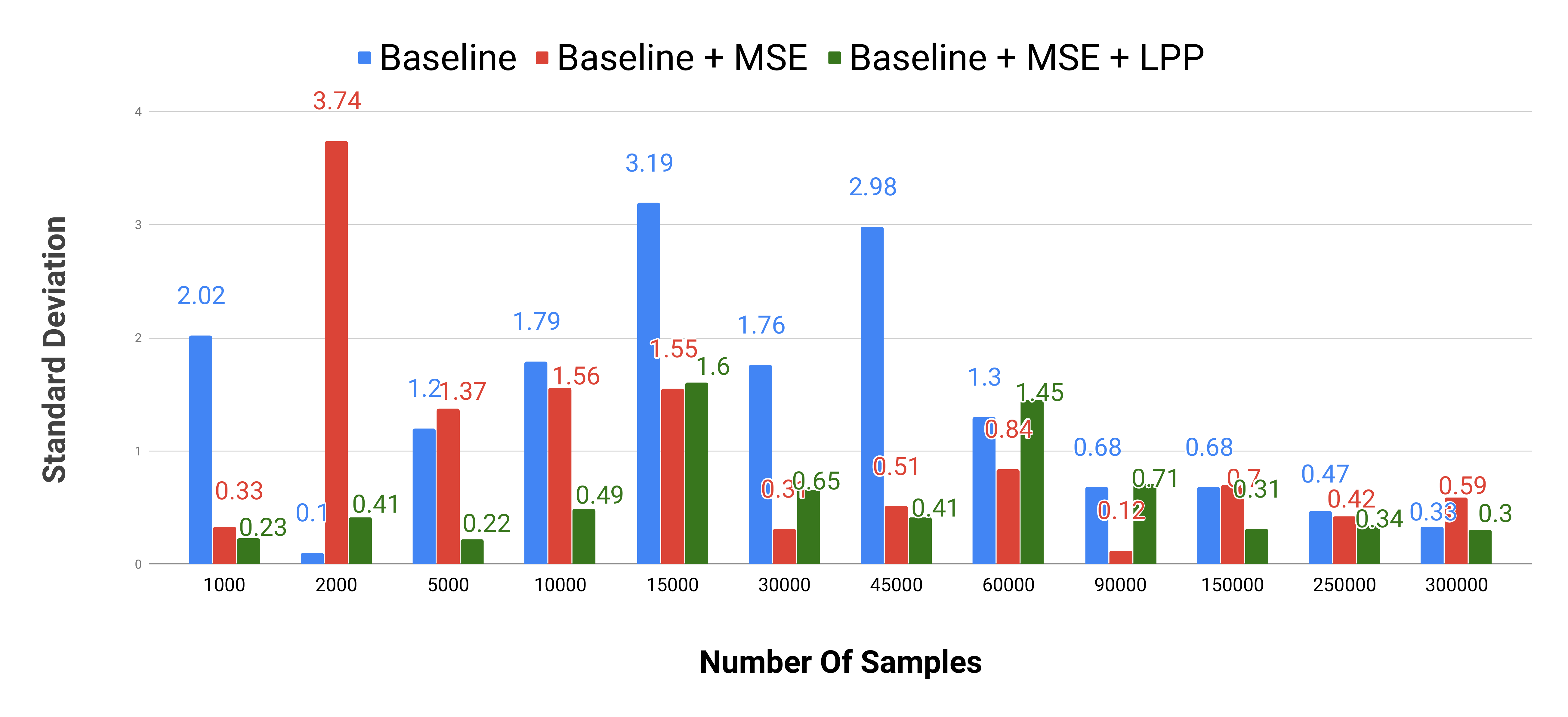}
\caption{\textbf{Standard deviation in accuracy of alignment on MNLI.} The graph shows standard deviation in accuracy when size of the training sample set differs (total: $300$K) for the baseline, baseline + MSE and baseline + MSE + LPL models: LPL yields more consistently optimal systems. (a) Standard deviation in accuracy when tested with in in-genre sentence pairs (matched MNLI) (b) Standard deviation in accuracy when tested with in out-of-genre sentence pairs (mismatched MNLI)}
    \label{fig:exp_mnli_stdev_perf}
\end{figure*}

\begin{table}[h]
    \centering
    \begin{tabular}{p{2.3cm} | p{4.5cm}}
        \hline
         \textbf{Manifold} & \textbf{Nearest Neighbors of ``Windows"}\Tstrut\\ 
        \hline
        Source ($M_1$) & nt4, 95/98/nt, nt/2000, nt/2000/xp, windows98\Tstrut\\
        Target ($M_2$) &  winzozz, mac, nt, osx, msdos\\
        Aligned ($f(M_1)$) &  winzozz, nt4, ntfs, mac, 95/98/nt, nt, osx, msdos\\
        \hline
    \end{tabular}
  \caption{Neighbors of the word ``windows'' in source domain (English), target domain (Italian) and the combined vector space with both English \& Italian vocabulary. The \textbf{Aligned} neighborhood contains a mix of the English and Italian words, not just the translation.}
  \label{tab:exp_neighbor_qual_analysis}
\end{table}

\Cref{tab:exp_neighbor_qual_analysis} shows the neighbors for the word ``windows" from the source embedding (English) and the target embedding (Italian). Compared to previous methods that look at explicit mapping of points between the two spaces, LPL tries to maintain the relations between words and their neighbors in the source domain while projecting them into the target domain. 

In this example, the word ``nt/2000" is not a part of the supervised pairs available and will not have an explicit projection in the target domain to be optimized without a locality preserving loss.

\section{Measuring Performance Consistency}
Figure \ref{fig:exp_stsb_stdev} and \ref{fig:exp_mnli_stdev_perf} chart the variation in evaluated performance on each model on tasks STS-B and MNLI when the size of the training dataset varies. In each case, a model optimized with the task specific loss has higher variation in performance than models that are optimized with additional losses MSE and LPL. The variation for baseline models are higher when the size of the dataset is small. Figure \ref{fig:exp_snli_stdev} shows the variation in test accuracy across runs on the SNLI dataset. The variation in accuracy is highest for the baseline model when the size of the dataset is small. For example, when the baseline is trained with $1000$ samples ($0.002$\% of the dataset), the variation in accuracy is $8.73$\%. Similarly, when the baseline is trained with $50$ samples ($0.01$\% of the dataset) in STS-B task, the variation in accuracy is $6.02$\% ($11.63$\% when sample size is $250$). The variation in accuracy reduces as the sample size increases. 

This occurs because when the subset of data randomly sampled, the quality of instances sampled has a large impact on the final performance of the model. Thus, measuring the consistency of the model's performance over multiple runs is a vital evaluation criteria (as much as the accuracy itself).

Hence, training the model with an alignment loss, i.e., with locality preservation ($\mathcal{L}_{mse}$ and $\mathcal{L}_{lpl}$), empirically guarantees that the model reaches near-optimal performance when the size of the supervised set is limited and that it has a narrow bound as compared to the baseline model trained without them. While performing these experiments, not only are the vocabularies randomly initialized, but also parameters too, making the model less dependent on how the training pairs are sampled from the dataset.

\section{Locally Linear Embedding}
\label{sec:app_lle_desc}
\begin{figure}[t]
    \centering
    \includegraphics[scale=0.29]{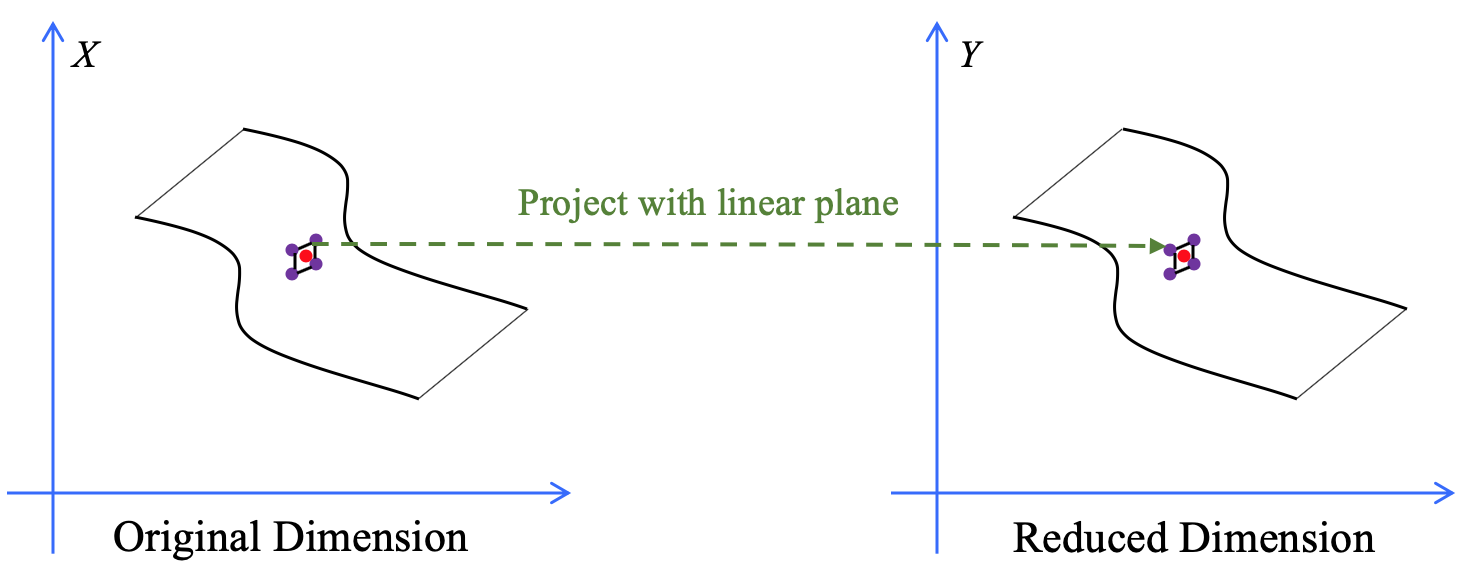}
\caption{\textbf{Locally Linear Embedding (LLE)}. The {\color{red} red}  point is a vector $X_i$ that is reconstructed by a plane of its neighbors in {\color{blue} blue}. This linear plane is functionally represented by $W\in\mathbb{R}^{n\times k}$ for $n$ points, each having $k$ neighbors.}
    \label{fig:lle_explanation}
\end{figure}

Our work is inspired from locally linear embedding (LLE) \citep{roweis2000nonlinear}. As discussed in \cref{sec:related_work_dr}, in LLE, the datapoints are assumed to have a linear relation with their neighbors (Figure \ref{fig:lle_explanation}). The reduced dimension projection of each vector is learned through a two step process: (a) Learn the linear relationship through a reconstruction loss (b) Use relation to learn low dimension representation. Assume each point in the manifold has $k$ neighbors $N_i$. The reconstruction loss is:
\begin{equation}
    \mathcal{L}_{\textrm{reconstruct}} = \sum_{i}{\norm{X_i - \sum_{j \in N_i}{W_{ij}X_{j}}}^2},
    \label{eq:alignment_lle_fs}
\end{equation}
where $X_i$ is the datapoint and the $X_j$ represents each neighbor. An additional constraint is imposed on the weights ($\sum_{ij} {W_{ij} = 1}$) to make the transform scale invariant. 

In \eqref{eq:alignment_lle_fs} the weights $W$ are an $N\times K$ matrix in a dataset of $N$ points (i.e., each point has its own weights). Given a learned $W$ from \eqref{eq:alignment_lle_fs}, we learn $Y_i$ (a projection for $X_i$) by minimizing the following:
\begin{equation}
    \sum_{i}{\norm{Y_i - \sum_{j \in N_i}{W_{ij}Y_{j}}}^2},
\label{eq:alignment_lle_ss}
\end{equation}
$Y_i$ is typically with reduced dimensions.





\FloatBarrier
\end{document}